\documentclass[journal]{IEEEtran}
\IEEEoverridecommandlockouts

\def\bibfiles{../} %

\usepackage{cite}
\usepackage{pgfplots}
\usepackage{csquotes}
\usepackage{amsmath,amssymb,amsfonts,amsthm}
\usepackage{algorithmic}
\usepackage{graphicx}
\usepackage{textcomp}
\usepackage{xcolor}
\usepackage{hyperref}       %
\usepackage{url}            %
\usepackage{booktabs}       %
\usepackage{verbatim}       %
\usepackage{ifthen}         %
\usepackage{tikz}
\pgfplotsset{compat=1.15}
\usepackage[vlined,boxed,commentsnumbered,linesnumbered,ruled]{algorithm2e}
\usepackage{comment}
\usepackage{balance}
\usepackage{bibunits}
\usepackage{xr}%
\usepackage{subfiles}
\usepackage{multicol}
\usepackage[capitalize]{cleveref} %
\usepackage[]{bibunits}

\newboolean{MAIN_REF} %
\setboolean{MAIN_REF}{false} %

\usepackage{nicefrac}  %
\usepackage{mathrsfs}  %
\usepackage{float}
\usepackage[caption=false,font=footnotesize,labelformat=simple]{subfig}
\crefname{equation}{\unskip}{\unskip}
\usetikzlibrary{shapes, patterns,decorations.text, decorations.pathreplacing}

\usepackage{bm}
\usepackage{enumitem}       %
\usepackage{dcolumn}        %
\usepackage{multirow}       %

\newtheorem{remark}{Remark}
\newtheorem{theorem}{Theorem}
\newtheorem{definition}{Definition}
\newtheorem{lemma}{Lemma}

\newtheorem{proposition}{Proposition}
\newtheorem{assumption}{Assumption}
\allowdisplaybreaks

\newcommand*{\Scale}[2][4]{\scalebox{#1}{\ensuremath{#2}}}%

\newcommand{\ssring}{\texttt{Skip-Ring}}
\newcommand{\ssrandring}{\texttt{Skip-Rand-Ring}}

\newcommand{\Reals}{\mathbb{R}}
\newcommand{\Naturals}{\mathbb{N}}   %
\newcommand{\set}[1]{\mathcal{#1}}
\newcommand{\tss}{t_\mathrm{skip}}
\newcommand{\vect}[1]{\mathbf{#1}} %
\newcommand{\vectg}[1]{{#1}} %
\renewcommand{\vect}[1]{\vectg{#1}} %
\DeclareMathOperator{\diag}{diag}
\newcommand{\dd}{\mathop{}\!\mathrm{d}}
\newcommand{\ee}{\mathrm{e}}
\newcommand{\eqdef}{\triangleq} %
\newcommand{\conv}[2]{#1 \mathrel{\star} #2}
\newcommand{\dW}{d_{\mathrm{W}_{\infty}}} %

\newcommand{\Prs}[1]{\Pr\left(#1\right)}

\newcommand{\Prscond}[2]{\Pr\left(#1 \kern0.1em\middle|\kern0.1em #2\right)}
\newcommand{\ePrscond}[2]{\Pr(#1 \kern0.1em|\kern0.1em #2)} 
\newcommand{\bigPrscond}[2]{\Pr\bigl(#1 \kern-0.1em \bigm| \kern-0.1em#2\bigr)}
\newcommand{\BigPrscond}[2]{\Pr\Bigl(#1 \kern-0.1em \Bigm| \kern-0.1em#2\Bigr)}
\newcommand{\biggPrscond}[2]{\Pr\biggl(#1 \kern-0.1em \biggm| \kern-0.1em#2\biggr)}
\newcommand{\BiggPrscond}[2]{\Pr\Biggl(#1 \kern-0.1em \Biggm| \kern-0.1em#2\Biggr)}

\newcommand{\Prv}[1]{\Pr\left[#1\right]}

\newcommand{\bigPrv}[1]{\Pr\bigl[#1\bigr]}

\newcommand{\Prvcond}[2]{\Pr\left[#1 \kern0.1em\middle|\kern0.1em #2\right]}
\newcommand{\ePrvcond}[2]{\Pr[#1 \kern0.1em|\kern0.1em #2]} 
\newcommand{\bigPrvcond}[2]{\Pr\bigl[#1 \kern-0.1em \bigm| \kern-0.1em#2\bigr]}
\newcommand{\BigPrvcond}[2]{\Pr\Bigl[#1 \kern-0.1em \Bigm| \kern-0.1em#2\Bigr]}
\newcommand{\biggPrvcond}[2]{\Pr\biggl[#1 \kern-0.1em \biggm| \kern-0.1em#2\biggr]}
\newcommand{\BiggPrvcond}[2]{\Pr\Biggl[#1 \kern-0.1em \Biggm| \kern-0.1em#2\Biggr]}

\newcommand{\EE}[2][]{%
	\ifthenelse{\equal{#1}{}}%
	{\mathbb{E} \left[ #2 \right]}%
	{\mathbb{E}_{#1} \left[ #2 \right]}%
}

\newcommand{\Exp}{\mathbb{E}}

\newcommand{\E}[2][]{\Exp_{#1}\left[#2\right]}

\newcommand{\bigE}[2][]{\Exp_{#1}\bigl[#2\bigr]}
\newcommand{\BigE}[2][]{\Exp_{#1}\Bigl[#2\Bigr]}
\newcommand{\biggE}[2][]{\Exp_{#1}\biggl[#2\biggr]}

\newcommand{\norm}[2][]{\left\|#2\right\|_{#1}}
\newcommand{\enorm}[2]{\|#2\|_{#1}}
\newcommand{\bignorm}[2][]{\bigl\|#2\bigr\|_{#1}}

\newcommand{\relD}{\mathop{}\!\mathscr{D}}         %
\newcommand{\relDf}[2]{\relD\left(#1 \kern0.1em\middle\|\kern0.1em #2\right)}
\newcommand{\erelDf}[2]{\relD(#1 \kern0.1em\|\kern0.1em #2)} 
\newcommand{\bigrelDf}[2]{\relD\bigl(#1 \kern-0.1em \bigm\| \kern-0.1em#2\bigr)}
\newcommand{\BigrelDf}[2]{\relD\Bigl(#1 \kern-0.1em \Bigm\| \kern-0.1em#2\Bigr)}
\newcommand{\biggrelDf}[2]{\relD\biggl(#1 \kern-0.1em \biggm\| \kern-0.1em#2\biggr)}
\newcommand{\BiggrelDf}[2]{\relD\Biggl(#1 \kern-0.1em \Biggm\| \kern-0.1em#2\Biggr)}
\newcommand{\RDiv}[3][]{\relD_{#1}\left(#2 \kern0.1em\middle\|\kern0.1em #3\right)}
\newcommand{\eRDiv}[3][]{\relD_{#1}(#2 \kern0.1em\|\kern0.1em #3)} 
\newcommand{\bigRDiv}[3][]{\relD_{#1}\bigl(#2 \kern-0.1em \bigm\| \kern-0.1em #3\bigr)}
\newcommand{\BigRDiv}[3][]{\relD_{#1}\Bigl(#2 \kern-0.1em \Bigm\| \kern-0.1em #3\Bigr)}
\newcommand{\biggRDiv}[3][]{\relD_{#1}\biggl(#2 \kern-0.1em \biggm\| \kern-0.1em #3\biggr)}
\newcommand{\BiggRDiv}[3][]{\relD_{#1}\Biggl(#2 \kern-0.1em \Biggm\| \kern-0.1em #3\Biggr)}

\newcommand{\sRDiv}[4][]{\relD_{#1}^{(#2)}\left(#3 \kern0.1em\middle\|\kern0.1em #4\right)}
\newcommand{\esRDiv}[4][]{\relD_{#1}^{(#2)}(#3 \kern0.1em\|\kern0.1em #4)} 
\newcommand{\bigsRDiv}[4][]{\relD_{#1}^{(#2)}\bigl(#3 \kern-0.1em \bigm\| \kern-0.1em #4\bigr)}
\newcommand{\BigsRDiv}[4][]{\relD_{#1}^{(#2)}\Bigl(#3 \kern-0.1em \Bigm\| \kern-0.1em #4\Bigr)}
\newcommand{\biggsRDiv}[4][]{\relD_{#1}^{(#2)}\biggl(#3 \kern-0.1em \biggm\| \kern-0.1em #4\biggr)}
\newcommand{\BiggsRDiv}[4][]{\relD_{#1}^{(#2)}\Biggl(#3 \kern-0.1em \Biggm\| \kern-0.1em #4\Biggr)}

\newcommand{\supRDiv}{\mathop{}\!\mathscr{R}}

\newcommand{\Normal}[2]{\mathcal{N}\left({#1},{#2}\right)} %
\newcommand{\eNormal}[2]{\mathcal{N}({#1},{#2})}

\newcommand{\veps}{\varepsilon}

\DeclareMathOperator*{\argmin}{arg\,min}

\definecolor{darkgreen}{rgb}{0, 0.5, 0}
\definecolor{SimulaBlue}{HTML}{445cea}

\newlength{\mywidth}
\newlength{\myheight}

\begin{document}
\title{Straggler-Resilient Differentially-Private \\[1mm] Decentralized  Learning}

\author{Yauhen~Yakimenka,~\IEEEmembership{Member,~IEEE},
  Chung-Wei~Weng,~%
  Hsuan-Yin~Lin,~\IEEEmembership{Senior Member,~IEEE},
  Eirik~Rosnes,~\IEEEmembership{Senior Member,~IEEE},
  and~J{\"o}rg~Kliewer,~\IEEEmembership{Fellow,~IEEE}
  \thanks{This work was supported by   the Experimental Infrastructure for Exploration of Exascale Computing (eX3), which is financially supported by the Research Council of Norway under Contract 270053. The work of Yauhen Yakimenka and Jörg Kliewer was  supported in part by U.S. NSF under Grant 1815322,  Grant 1908756, and Grant 2107370. This paper was presented in part at the IEEE Information Theory Workshop (ITW), Mumbai, India, November 2022 \cite{YakimenkaWengLinRosnesKliewer22_1}.}
  \thanks{Y.~Yakimenka and J.~Kliewer are with Helen and John C. Hartmann Department of Electrical and Computer Engineering, New Jersey Institute of Technology, Newark, New Jersey 07102, USA (e-mail: \{yauhen.yakimenka, jkliewer\}@njit.edu).}
  \thanks{C.-W.~Weng, H.-Y.~Lin, and E.~Rosnes are with Simula UiB, N-5006 Bergen, Norway  (e-mail: \{chungwei, lin, eirikrosnes\}@simula.no).}
}

\maketitle

\setlength{\mywidth}{0.46\columnwidth}
\setlength{\myheight}{0.4\columnwidth}

\begin{abstract}
We consider the straggler problem in decentralized learning over a logical ring while  preserving user data privacy. Especially, we extend the recently proposed framework of differential privacy (DP) amplification by decentralization by Cyffers and Bellet to include overall training latency\textemdash comprising both computation and communication latency. Analytical results on both the convergence speed and the DP level are derived for both a skipping scheme (which ignores the stragglers  after a timeout) and a baseline scheme that waits for each node to finish before the training continues. A trade-off between overall training latency, accuracy, and privacy, parameterized by the timeout of the skipping scheme, is identified and empirically validated for logistic regression on a real-world dataset and for image classification using the MNIST and CIFAR-$10$ datasets. 
 \end{abstract}


\section{Introduction}
In distributed learning,  a finite-sum optimization problem is solved across multiple nodes without exchanging the local datasets directly, thus improving user data privacy and  reducing the communication cost. A popular instance of distributed learning is federated learning~\cite{McMahanMooreRamageHampsonArcas17_1,Konecy-etal16_1,LiSahuTalwalkarSmith20_1}
 in which there is a single central server  coordinating the training process. On the other hand, in fully decentralized learning, see, e.g.,~\cite{Lian-etal17_1,XiongYanSinghLi21_1sub}, there is no such coordinating central server\textemdash the nodes maintain a local estimate of the optimal model and iteratively update it by averaging estimates obtained from neighbors  corrected on the basis of their local datasets.
There are two modes of operation\textemdash sequential and parallel\textemdash and  
theoretical studies show that the  physical communication topology has a strong impact on the number of epochs needed to converge \cite{NegliaXuTowsleyCalbi20_1}.

It is well-known by now that the computed partial (sub)gradients can leak information on the local datasets \cite{FredriksonJhaRistenpart15_1}. %
In order to circumvent this, a  carefully selected noise term can be added to the computed partial (sub)gradients before they are transmitted to other nodes, referred to as local differential privacy (LDP)  \cite{GeyerKleinNabi17_1sub,Wei-etal20_1}. In fully decentralized learning, nodes have only a local view of the system. Hence,  Cyffers and Bellet \cite{CyffersBellet22_1} recently proposed a novel relaxation of LDP, referred to as network DP (NDP),  to naturally capture this. Furthermore, they showed that the  privacy-utility trade-off under  NDP can be significantly improved upon compared to what is achievable under LDP, illustrating  that formal privacy gains can be obtained from full decentralization, complementing previous notions of ``amplifying'' the privacy   by shuffling, subsampling, and iteration~\cite{BalleBartheGaboardi18_1,FeldmanMironovTalwarThakurta18_1,ErlingssonFeldman-etal19_1,FeldmanMcMillanTalwar22_1}.  
Recently,  the work in \cite{CyffersBellet22_1} was extended to a parallel approach  that alternates between local gradient descent steps for all nodes in  parallel and subsequent gossip averaging \cite{CyffersEvenBelletMassoulie22_1}. Accordingly, the NDP concept was relaxed to capture that the privacy leakage from a node to another node may depend on their distance in the graph. It was shown in \cite{CyffersEvenBelletMassoulie22_1}  that privacy amplification can be achieved as for the sequential approach in \cite{CyffersBellet22_1}.
Differentially-private fully decentralized learning has also been considered in several other previous works, see, e.g., \cite{XiongYanSinghLi21_1sub,ShowkatbakhshKarakusDiggavi19_1sub,JinHeDai18_1sub}. In the federated learning case, there are numerous works that consider user privacy, e.g., both from a DP perspective (see, e.g., \cite{Wei-etal20_1}) and from an information-theoretic secure aggregation perspective (see, e.g., \cite{Bonawitz-etal17_1,KadheRajaramanKoyluogluRamchandran20_1sub,So-etal22_1,SoGulerAvestimehr21_2,SchlegelKumarRosnesGraellAmat23_1}).

The problem of \emph{straggling} nodes, i.e., nodes that take a long time to finish their tasks due to random phenomena such as processes running in the background and memory access, has been broadly studied in the literature. The \emph{ignoring-stragglers strategy}, i.e., ignoring results from the slowest nodes, see, e.g., \cite{ReisizadehTaheriMokhtariHassaniPedarsani19_1,XiongYanSinghLi21_1sub}, is simple and popular, but can lead to convergence to a local optimum when the data is  heterogeneous~\cite{CharlesKonecy20_1sub,MitraJaafarPappasHassani21_1}. 
Coded computing methods \cite{LeeLamPedarsaniPapailiopoulosRamchandran18_1,LiAvestimehr20_1,Yu20_1} is an alternative to provide resiliency against straggling nodes, and the key idea is to add redundancy to the computation through an error-correcting code.  %
The coded computing literature has considered several different computing tasks, e.g., vector-matrix multiplication \cite{LeeLamPedarsaniPapailiopoulosRamchandran18_1,SeverinsonGraellAmatRosnes19_1,DuttaCadambeGrover19_1}, (secure) distributed matrix-matrix multiplication \cite{YuMaddah-AliAvestimehr17_1,Dutta-etal20_1,YuAvestimehr21_1,ChangTandon18_1,KakarEbadifarSezgin19_1,YangLee19_1,AliasgariSimeoneKliewer19_1,DOliveiraElRouayhebKarpuk20_1,YuMaddah-AliAvestimehr20_1,MitalLingGunduz22_1}, and more general distributed optimization and nonlinear computation problems \cite{TandonLeiDimakisKarampatziakis17_1,KarakusSunDiggaviYin17_1,YangPedarsaniAvestimehr19_1,YuLiRavivKalanSoltanolkotabiAvestimehr19_1,BitarWoottersElRouayheb20_1,WangDuursma20_1sub,YangAvestimehr21_1,KosaianRashmiVenkataraman20_1}. For matrix-matrix multiplication, the state-of-the-art for straggler mitigation is achieved by the combination of the results in  \cite{WangDuursma20_1sub} and \cite{YuMaddah-AliAvestimehr20_1}.

In this work, we study the impact of stragglers and user data privacy  in decentralized training. In particular, we assume an underlying physical full mesh topology, i.e., all nodes can physically communicate with each other, but sequential training along  a logical ring on top of the physical topology where each node communicates a token only with its immediate neighbors upstream and downstream. In sequential training, nodes do not need to be active during the whole training period, which makes it suitable for scenarios where the nodes have limited resources, and therefore remain dormant unless they are triggered to do an update. See also \cite{ElkordyPrakashAvestimehr22_1,wang2021efficient} for further motivation for this scenario. For this setting, we extend the recently proposed framework of privacy amplification by decentralization by Cyffers and Bellet\cite{CyffersBellet22_1} to include the overall latency\textemdash comprising both computation and communication latency\textemdash %
under stochastic gradient descent. Our main contributions are summarized as follows.

\begin{itemize}

\item We study  a skipping scheme (which ignores the stragglers after a timeout) and a baseline scheme that waits for each node to finish its computation before the training continues, for  a fixed and a randomized ring topology, and  derive analytical results on the convergence behavior (see \cref{thm:convergence}) and the DP level (see \cref{thm:ss_ring_all_noise,thm:ss_rand_ring_all_noise}), revealing a trade-off  parameterized by the timeout of the skipping scheme. We show that the asymptotic convergence rate is equal to that of \cite[Thm.~2]{ShamirZhang13_1}. We note that the presented proofs in Appendices~\ref{app:A} and \ref{app:AppB} require several nontrivial steps which can not be found in previous work, e.g., the asymptotic convergence analysis in %
Appendix~\ref{sec:asympt-conv-rate} in the supplementary material and the adaption to a decreasing learning rate in Appendix~\ref{app:AppB}. See also the first paragraph of \cref{sec:convergence-analysis}. Moreover, we emphasize again that this work studies the effect of stragglers, which by itself is novel for the considered scenario.

\item The optimal timeout that minimizes the time between two consecutive updates of the  token is determined, showing that skipping is beneficial for faster convergence for certain popular  computational delay models considered in the literature (see \cref{prop:optimal_tss,sec:numerical_results_regression}).

\item We show that randomizing the processing order of nodes on the ring yields an improvement in both convergence
behavior and privacy in the long run  (see \cref{sec:numerical_results_theory}), although the error and the privacy leakage level show the same order-wise asymptotic behavior in the number of update steps with and without randomization (see \cref{rem:1}). This is in particular prominent for a larger number of nodes due to the increased effect of privacy amplification.
\end{itemize}

Finally, we present  extensive empirical results  for both logistic regression on a binarized version of the UCI housing dataset~\cite{OpenMLdata14_1} and for image classification using both the MNIST~\cite{LecunBottouBengioHaffner98_1} and CIFAR-$10$~\cite{Krizhevsky09_1} datasets to validate our theoretical findings. We also compare with the  parallel approach from \cite{CyffersEvenBelletMassoulie22_1} and to a centralized federated learning approach.\footnote{Compared to the conference version \cite{YakimenkaWengLinRosnesKliewer22_1}, we provide  a \emph{complete} exposition that includes all  technical proofs, as well as new asymptotic results, in addition to significantly extended numerical results. Missing proofs (including the proof of \cref{rem:1}) can be found in Appendices~\ref{app:proofs} and \ref{sec:asympt-conv-rate} in the supplementary material. The code for this work is available at \url{https://github.com/Simula-UiB/SRDPDL_JSAIT24}.}

\section{Preliminaries}

\subsection{Notation}
\label{sec:notation}
We use uppercase and lowercase letters for random variables (RVs) and their realization (both scalars and vectors), respectively, and italics for sets, e.g., \(X\), \(x\),  and \(\mathcal{X}\)  represent a RV, a scalar/vector, and a set, respectively.  
An exception to this rule is $\tau$ which denotes  the   model description, also referred to as the token. %
Matrices are  denoted by uppercase letters, their distinction from RVs will be clear from the context.
Vectors are represented as row vectors and the transpose of a vector or a matrix is denoted by $(\cdot)^\top$.   The expectation of a RV $X$ is denoted by $\EE{X}$. We define $[n] \triangleq \{1,2,\dotsc, n\}$, %
while $\mathbb{N}$ denotes  the set of natural numbers and $\Reals$  the set of real numbers. The (sub)gradient of a function $f(x)$ is denoted by $\nabla f(x)$, while the $\ell_p$-norm of a length-$n$ vector $x=(x_1,\ldots,x_n) \in \Reals^n$ is denoted by $\lVert x \rVert_p = \left(\sum_{i=1}^n |x_i|^p \right)^{\nicefrac{1}{p}}$, where $|\cdot|$ denotes absolute value. The base of the natural logarithm is denoted by ${\mathrm e}$, while $\log$ denotes  natural logarithm. 
$\Normal{\mu}{\sigma^2 I_d}$ denotes the $d$-dimensional Gaussian (uncorrelated) distribution with mean $\mu$ and standard deviation $\sigma$ of each component, where $I_d$ is the identity matrix of size $d$. 
$X \sim \set P$ denotes that $X$ is distributed according to the distribution $\set P$, while $x \sim \set P$ denotes a sample  $x$ taken from  $\set P$. %
We denote by $\set D \sim_u \set D'$ the fact that datasets $\set D = \cup_{v \in \set V} \set D_v$ and $\set D' = \cup_{v \in \set V} \set D_v'$ are the same except perhaps for the dataset of  user $u$, i.e., $\set D_v = \set D_v'$ for all $v \neq u$, where $\set V$ is some set of users. Standard order notation $O(\cdot)$ is used for asymptotic results.

\subsection{Definitions and Assumptions}

\begin{definition}[$k$-Lipschitz continuity]
A function $f: \mathcal{W} \rightarrow \mathbb{R}$ is $k$-\emph{Lipschitz continuous} over the convex domain $\mathcal{W} \subseteq \mathbb{R}^d$ if $| f(\vect{w})-f(\vect{w}')| \leq k \norm[2]{\vect{w}-\vect{w}'}$ for all $\vect{w},\vect{w}' \in \mathcal{W}$.
\end{definition}

\begin{definition}[$\beta$-smoothness]
A function $f\colon \mathcal{W} \rightarrow \mathbb{R}$ is $\beta$-\emph{smooth}  over the convex domain $\mathcal{W} \subseteq \mathbb{R}^d$ if $\norm[2]{\nabla f(\vect{w})- \nabla f(\vect{w}')} \leq \beta \norm[2]{\vect{w}-\vect{w}'}$ for all $\vect{w},\vect{w}' \in \mathcal{W}$.
\end{definition}

\begin{assumption} \label{ass:Lipschitz}
$f_v(\tau;\cdot)$, $v \in \mathcal{V}$, is $k$-Lipschitz continuous %
 and convex  in its first argument. 
\end{assumption}

\begin{assumption} \label{ass:smooth}
$f_v$, $v \in \set{V}$,  is $\beta$-smooth.
\end{assumption}

\vspace{-3mm}
\subsection{System Model}
Consider a decentralized network of $n$ honest-but-curious nodes (users) $\set V = \{v_1, \dotsc, v_n\}$ with a decentralized dataset $\set D = \cup_{v \in \set V} \set D_v$ where $\set D_v = \bigl\{(x^{(v)}_i, y^{(v)}_i)\bigr\}_{i=1}^\kappa$, $(x^{(v)}_i,y_i^{(v)}) \in \set{R} \subseteq \Reals^{d_\mathrm{x}}\times \Reals^{d_\mathrm{y}}$, for some set $\mathcal{R}$ and $d_\mathrm{x},d_\mathrm{y},{\kappa} \in \mathbb{N}$, %
is the private dataset of  node $v \in \set V$. %

The nodes want to compute some function together based on their datasets but want to keep their datasets private. For that, they employ a decentralized protocol where a token $\tau \in \mathcal{W}$, for some convex set $\mathcal{W} \subseteq \mathbb{R}^d$, travels between the nodes according to some predefined (but potentially randomized) path. %
When receiving the token the $r$-th time and the global time is $h$, the node $v$ updates it as
$\tau \leftarrow g^{(v)}_r(\tau;\, \mathtt{state}_v(h))$,
and sends it further. Here, $\mathtt{state}_v(h)$ encapsulates all the information available to the node $v$ at time $h$, e.g., the available data points and the results of previous calculations. It can also include some source of randomness. We assume that the computation in each node $v$ during the $r$-th visit of the token  takes random time $T^{(v)}_r$. Hence,  the computation of $g^{(v)}_r(\cdot, \cdot)$ takes time at most $T^{(v)}_r$ as the token may be updated before the entire computation is finished.\footnote{The RVs $T^{(v)}_r$ are assumed to be independent and identically distributed (i.i.d.) which is in accordance with the literature, where typically stragglers are generated uniformly at random, except for a few works, e.g., \cite{YangPedarsaniAvestimehr19_1,SeverinsonRosnesElRouayhebGraellAmat23_1} that consider a model where nodes tend to remain stragglers for a long time, violating the i.i.d. assumption on the RVs $T^{(v)}_r$.} %
We consider a model where  $T^{(v)}_r$ is comprised of a deterministic constant part (the time it takes for  an actual computation) and a random part.  Also, we assume that communication between any two nodes is noiseless and takes constant time $\chi$, and hence the constant part of the computation time can be set to zero. %
At the end of the protocol, the token $\tau$ is distributed among  the nodes, which allows for calculating the desired result. This final distribution takes constant overhead time and is therefore ignored.

For a decentralized protocol $\set A$, we denote by $\set A(\set D)$ the (random) transcript of all messages sent or received by all the users, i.e., $\set A(\set D)$ are all the triples $(u,w,v)$, if $u\in \set V$ sent a message with content $w$ to  $v \in \set V$. %
However, due to the decentralized nature of $\set A$, the user $v$ only has access to the subset of $\set A(\set D)$ consisting of the messages she sent or received, and we denote this view by $\set O_v(\set A(\set D)) = \{(u,w,u') \in \set A(\set D) : \text{$u=v$ or $u'=v$}\}$. %
Let $\Omega$ denote the set of all possible views, i.e., $\set O_v(\set A(\set D)) \in \Omega$ for all possible parameters and realizations.

\vspace{-0.1ex}
\subsection{Network Differential Privacy}
\label{sec:network-DP}

We accept the notion of NDP introduced in \cite{CyffersBellet22_1}. 
\begin{definition}[NDP \cite{CyffersBellet22_1}]
  A protocol $\set A$ satisfies $(\varepsilon,\delta)$-NDP if for all pairs of distinct users $u,v \in \set V$, all pairs of neighboring datasets $\set D \sim_u \set D'$, and any $\set S \subseteq \Omega$, we have
  \begin{IEEEeqnarray*}{c}
    \Pr[\set O_v(\set A(\set D)) \in \set S] \leq {\mathrm{e}}^\varepsilon \Pr[\set O_v(\set A(\set D')) \in \set S]  + \delta,
  \end{IEEEeqnarray*}
  where the notion of neighboring datasets $\set D \sim_u \set D'$ is defined in \cref{sec:notation}.
\end{definition}

NDP  measures  how much the information collected by node $v$ depends on the dataset of node $u$. In the special case that all nodes can observe all messages, 
i.e., $\set O_v$ is the identity map, NDP boils down to conventional LDP \cite{DuchiJordanWainwright13_1}. %
When processing information in a decentralized manner with no central  coordinating entity, and when there is no third party (on top of the topology) observing all messages sent, NDP is a more natural privacy measure than DP or LDP. 

\section{Empirical Risk Minimization}

In this section, we consider the empirical risk minimization problem
\begin{IEEEeqnarray}{c}
  \label{eq:min-problem}
  \tau^* = \argmin_{\tau \in \mathcal W \subseteq \Reals^d}\, \left[ f(\tau;\set D) \triangleq \frac 1n \sum_{v \in \set V}  f_v(\tau;\mathcal{D}_v) \right], %
\end{IEEEeqnarray}
where $f_v(\tau;\cdot)$ is $k$-Lipschitz continuous %
and convex 
in its first argument (see \cref{ass:Lipschitz}).

\subsection{Skipping Scheme}
\label{sec:ss}

We suggest the following protocol inspired by projected noisy stochastic gradient descent  to solve \eqref{eq:min-problem}. The token $\tau$ keeps the current estimate of the optimal point $\tau^*$ and follows a possibly randomized path  over the available nodes  $\mathcal{V}$. 
To speed up the process, the token waits up to a threshold time $\tss$ and, if the computation has not finished by that time, the token is forwarded further without an update.\footnote{In practice, acknowledgments can identify straggling nodes: if the token is sent to the next node in line and not acknowledged within a threshold time, it is forwarded to the next node in line, 
etc.} In our notation, it means that the calculation in each  node $v$ is
\begin{align}
		&g^{(v)}_r(\tau; \mathtt{state}_v(h)) \notag \\
		&=   \begin{cases}
			\Pi_{\set W} \left(\tau -\eta_h \left( \nabla f_v(\tau; \set D_v)+N_h\right)\right) & \text{if $T_r^{(v)} \leq \tss$}, \label{eq:g_update}\\
			\tau & \text{otherwise},
		\end{cases}
\end{align}
where $\eta_h$ is the step size (learning rate), $\Pi_\mathcal{W}$ denotes the Euclidean projection onto the set $\mathcal{W}$, %
and $N_h$ is  noise with zero mean and standard deviation $\sigma_h$. %
The noise $N_h$ is added in order to protect the privacy of the local datasets, and the standard deviation $\sigma_h$ is chosen so a certain level of NDP is ensured.\footnote{The noise follows $\Normal{0}{\sigma_h^2}$, where $\sigma_h = \frac{k \sqrt{8 \log (\nicefrac{1.25}{\delta})}}{\varepsilon}$, %
$\varepsilon > 0$, and $0 < \delta < 1$.} In this work, we consider %
the gamma distribution (including the exponential distribution) and the Pareto type II (also known as Lomax) distribution for $T_r^{(v)}$, which are well-established models in the literature, see, e.g., \cite{DuttaCadambeGrover17_1,SeverinsonRosnesElRouayhebGraellAmat23_1,NegliaCalbiTowsleyVardoyan19_1}. Since we assume that the RVs $T_r^{(v)}$ are i.i.d., we simplify the notation in the following by letting $T \equiv T_r^{(v)}$.

\begin{figure}[t]
	\centering
	\subfloat[\ssring.]{%
		 \includegraphics[width=0.47\columnwidth]{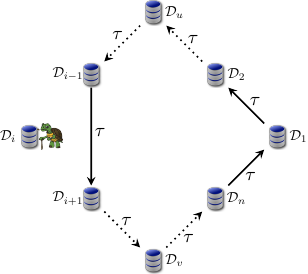} %
		 }
	\hfill
	\subfloat[\ssrandring.]{%
		 \includegraphics[width=0.47\columnwidth]{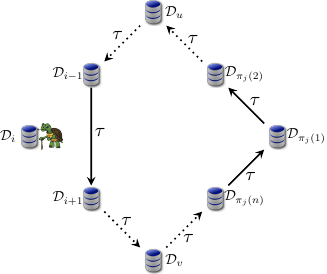}
		 }
	\caption{Illustrating the $j$-th round in which node $v_i$ is a straggler.} 
	\label{fig:systemmodel}
	\vspace{-3mm}
\end{figure}

The algorithm stops when a predefined convergence requirement is fulfilled. We refer to the algorithm detailed above as the skipping scheme with parameter $\tss$, which can be optimized in order to reduce either  the convergence time and/or the privacy leakage. In the special case of $\tss=\infty$, it reduces to a scheme for which the token always waits. We denote by $p = \Pr[T > \tss] %
$ the probability of skipping a node. The formal algorithm is given  in \cref{alg:ss}, where the output $\ell$ denotes its execution latency and $\tau_{h_{\max}}$ the final value of the token after $h_{\max}$ steps.

\begin{algorithm}[t]
	\KwIn{Datasets $\set D_v$ and $k$-Lipschitz continuous %
	convex functions $f_v: \set W \times \set R^\kappa \rightarrow \Reals$, $v \in \set V$, in the first argument, %
	 noise standard deviation sequence $(\sigma_1,\ldots,\sigma_{h_{\max}})$, node path sequence $(v^{(1)},\ldots,v^{(h_{\max})})$, learning rate parameter $\zeta$, skipping parameter  $\tss$, number of steps $h_{\max}$, and communication latency $\chi$}
	\KwOut{$(\tau_{h_{\max}},\ell)$}
        $\tau_0 \leftarrow 0$, \,
        $\ell \leftarrow 0$, \, $c\leftarrow1$\\
        $\set P \leftarrow \text{Comp. lat. model}$ (gamma or Pareto type II)\\
\For{$h \in [h_{\max}]$}{
$t\sim \set P$\\
\If{$t \leq \tss$}{
$\eta_h \leftarrow \zeta/\sqrt{c}$\\
$\tau_{h} \leftarrow \Pi_\mathcal{W} \left( \tau_{h-1} - \eta_h \left( \nabla  f_{v^{(h)}}(\tau_{h-1}; \set D_{v^{(h)}}) + N_h \right) \right)$, where $N_h\sim\mathcal{N}(0,\sigma_h^2 I_d)$\label{alg:ss:token-update}\\
$\ell \leftarrow \ell +  \chi+t$, \, 
$c \leftarrow c+1$
}
\Else{$\tau_{h}  \leftarrow \tau_{h-1}$, \,
$\ell \leftarrow \ell +  \chi+\tss$}}
\KwRet{($\tau_{h_{\max}},\ell)$}
	\caption{Skipping Scheme}
	\label{alg:ss}
\end{algorithm}

We  use \cref{alg:ss} in two special cases as outlined below and illustrated in \cref{fig:systemmodel}. For both schemes, the noise variance is fixed  throughout the algorithm, i.e., $\sigma_h = \sigma$, $\forall\, h$, and we assume, for simplicity, that $h_{\max}$ is a multiple of $n$ in the rest of the paper. %

\begin{itemize}
\item First, we consider an update schedule in which the nodes in $\set V$ are processed along a logical ring, i.e., the node path sequence of \cref{alg:ss} is $(v^{(1)},\ldots,v^{(h_{\max})})=((v_1,\ldots,v_n),(v_1,\ldots,v_n),\ldots,(v_1,\ldots,v_n))$.  
  The corresponding scheme is denoted by \ssring.
\item Second, we consider a \emph{randomized} version of the logical ring, denoted by \ssrandring. %
Each round over the ring can be seen as a random walk on the set of nodes, but without replacement. For each round, the random walk procedure is restarted. Hence, the node path sequence becomes $(v^{(1)},\ldots,v^{(h_{\max})})= ((v_{\pi_1(1)},\ldots,v_{\pi_1(n)}), (v_{\pi_2(1)},\ldots,v_{\pi_2(n)}),  \ldots,$ $(v_{\pi_{h_{\max}/n}(1)},\ldots,v_{\pi_{h_{\max}/n}(n)}))$ 
  where %
  $\pi_1, \ldots, \pi_{h_{\max}/n}$  are independent random permutations over  $[n]$.
\end{itemize}

As a final remark, we mention here that results on the computation and communication latency for the skipping scheme in \cref{alg:ss} will be presented later in \cref{sec:comp-comm-latency}.

\section{Convergence Analysis}
\label{sec:convergence-analysis}
Here, we provide a convergence result for the two considered schemes by adapting the classical convergence result of \cite[Thm.~2]{ShamirZhang13_1} to decentralized learning where nodes are processed according to a Markov chain
and for which the (sub)gradient estimate in each step is biased, but \emph{converges to unbiased} exponentially fast, which are the main two new technicalities of the proof.\footnote{There are several previous works that provide convergence results for   Markov chain (noisy) stochastic gradient descent, e.g., \cite{SunSunYin18_1,AyacheElRouayheb21_1}. However, all of these works require that  $\sigma_h$  decays to zero with  $h$, which means a significantly higher leakage of private data. %
}
Additionally, the number of token updates is random (depending on the skipping probability), and we need to average over it.  Note that, as in \cite[Thm.~2]{ShamirZhang13_1}, $f_v$, $v \in \set{V}$, is not required to be $\beta$-smooth 
or even $k$-Lipschitz continuous, as we only need the (sub)gradients to be bounded (which follows from $k$-Lipschitzness), and also that our result  provides a guarantee on the performance of the last update of the token instead of for the average of all token values.

\begin{theorem}
  \label{thm:convergence}
  Under \cref{ass:Lipschitz}, if the diameter of $\mathcal W$ is $d_{\mathcal W}$, %
  the expected difference between the minimum value $f(\tau^*;\cdot)$ and that from \cref{alg:ss} with an arbitrary  learning rate parameter $\zeta > 0$
 after $h_{\max}$ steps is bounded as
\begin{IEEEeqnarray*}{rCl}
    \EE{f(\tau_{h_{\max}};\cdot) - f(\tau^*;\cdot)}&\leq&\sum_{h=0}^{h_{\max}} \binom{h_{\max}}{h} (1-p)^h p^{h_{\max}-h} e_h
    \IEEEeqnarraynumspace
     \\
     & = &  O\left( \frac{\log (h_{\max})}{ \sqrt{h_{\max}}} \right),
\end{IEEEeqnarray*}
where $\forall\, h > 0$,
\begin{multline*}
	e_{h} \eqdef \frac{(d_{\mathcal W}^2 + \zeta^2(k^2 + d \sigma^2))(2+\log (h+1))}{\zeta\sqrt{h+1}}
	\\[-3pt]
	\quad\,\,\, +\,\,  d_{\set W} k \sqrt n \Biggl( \frac{1}{h+1} \sum_{i=1}^{h+1} |\lambda_1|^i + \sum_{j=1}^{h} \frac{1}{j(j+1)} \sum_{i = h +1- j}^{h+1} |\lambda_1|^i \Biggr)
\end{multline*} 
and $e_0 \eqdef d_{\set W} k$, 
$|\lambda_1| = \frac{1-p}{\sqrt{(1+p^2) - 2p\cos(\frac{2\pi}{n})}}$ and $0 < p < 1$
for \ssring, while $\lambda_1 \eqdef 0$ and $0 \leq p < 1$ for \ssrandring. %
\end{theorem}
	
\begin{IEEEproof}
See Appendices~\ref{app:A}  and \ref{sec:asympt-conv-rate} in the supplementary material for  the finite and asymptotic results, respectively.
\end{IEEEproof}
Note that the asymptotic convergence rate is the same as that of \cite[Thm.~2]{ShamirZhang13_1}, while being a $\log(h_{\max})$-factor worse compared to \cite[Thm.~1]{SunSunYin18_1}. The latter is due to 1) the assumption %
that $\sigma_h$ %
decays to zero with  $h$ \cite[Eq.~(16)]{SunSunYin18_1}, and 2) that convergence there is proved for the running average of the token.

Interestingly, 
the asymptotic behavior of the bound in \cref{thm:convergence} is the same for both $\lambda_1=0$ and $\lambda_1 > 0$. Hence, a biased (sub)gradient estimate that converges to unbiased exponentially fast does not influence the asymptotic convergence rate. Moreover, in \cref{thm:convergence}, we do not allow for $p=0$ in the \ssring\ scheme as in this case the stochastic (sub)gradient is biased, even asymptotically, and hence   a different proof technique is required. The asymptotic convergence rate in this special corner case is left open. Note that the proof of \cite[Thm.~2]{ShamirZhang13_1} cannot be adapted to this scenario as it requires an unbiased stochastic (sub)gradient.

\begin{remark}
For the uniform random walk scheme considered in \cite{CyffersBellet22_1}, the marginal distribution of visited nodes at each step is uniform, as it is with \ssrandring. Thus, 
the proof of
\cref{thm:convergence} applies to both these schemes with $\lambda_1 \eqdef 0$ and $0 \leq p < 1$.
\end{remark}

\section{Privacy Analysis} \label{sec:privacy}

In this section, we present results on the privacy leakage level of the skipping scheme for both updating schedules of the token outlined in \cref{sec:ss}, i.e., for both a fixed and a randomized logical ring on the set of nodes $\set V$. We highlight here that compared to \cite{CyffersBellet22_1}, that only considers  a constant learning rate and also a different randomized path (and no fixed path), our results apply to a decreasing learning rate of the form $\eta_h = \zeta / \sqrt{h}$ (as specified  in \cref{alg:ss}).

The full proof, which can be found in Appendix~\ref{app:AppB}, revolves around upper bounding the R{\'e}nyi divergence between $\set O_v(\set A(\set D))$ and $\set O_v(\set A(\set D'))$, $\set D \sim_u \set D'$, for any distinct pair of users $u,v$, using tools (including a composition theorem for R{\'e}nyi DP  (RDP) \cite[Prop.~1]{Mironov17_1}) from the framework of privacy amplification by iteration\cite{FeldmanMironovTalwarThakurta18_1}. %
The resulting bound can be transformed into a bound on NDP using \cite[Prop.~3]{Mironov17_1} and further optimized.
Allowing for a decreasing learning rate constitutes the main technical contribution of the proof.%

\begin{theorem}
  \label{thm:ss_ring_all_noise}
  Let $\varepsilon > 0$  
  and $0 < \delta < 1$. Then, under Assumptions~\ref{ass:Lipschitz} and \ref{ass:smooth}, the \ssring\   scheme on a ring with $n$ nodes %
  and with  learning rate parameter
  $0 < \zeta \leq \nicefrac{2}{\beta}$ achieves $(\varepsilon_{\mathrm{skip}},\delta+\delta')$-NDP for all $\delta' \in (0,1]$ with
  \begin{IEEEeqnarray*}{c}
    \varepsilon_{\mathrm{skip}}=\varepsilon \frac{\sqrt{\tilde{h} \log(\nicefrac{1}{\delta})} }{\sqrt{\log(\nicefrac{1.25}{\delta})}}   + \frac{ \varepsilon^2 \tilde{h}}{ 4 \log(\nicefrac{1.25}{\delta})},  
  \end{IEEEeqnarray*}
where 
  $\tilde{h}\triangleq\Bigl\lceil \nicefrac{h_{\max}(1-p)}{n}+ \sqrt{\nicefrac{3h_{\max}(1-p)}{n}   \log \left( {\nicefrac{1}{\delta'}} \right)}\Bigr\rceil$ %
and $0 \leq p < 1$ is the probability of skipping a node.
\end{theorem}

The following theorem characterizes the privacy leakage level $\varepsilon_{\mathrm{skip}}$ of the \ssrandring\ scheme.

\begin{theorem}
  \label{thm:ss_rand_ring_all_noise}
  Let $\varepsilon > 0$ 
  and $0 < \delta < 1$. Then, under Assumptions~\ref{ass:Lipschitz} and \ref{ass:smooth},  the \ssrandring\  scheme on a ring with $n$ nodes %
  and with learning rate parameter  $0 < \zeta \leq \nicefrac{2}{\beta}$
   achieves $(\varepsilon_{\mathrm{skip}},\delta+\delta')$-NDP for all $\delta' \in (0,1]$ with
  \begin{IEEEeqnarray*}{c}
    \varepsilon_{\mathrm{skip}} = \frac{\varepsilon^2 a  \alpha}{2 \log (\nicefrac{1.25}{\delta})}   + \frac{\log (\nicefrac{1}{\delta})}{\alpha-1}, 
      \end{IEEEeqnarray*}
  where
  \vspace{-1ex}
\begin{IEEEeqnarray*}{rCl}
        a& \triangleq & \frac{1}{n-1} \sum_{r=0}^{\tilde{h}-1} \sum_{d=1}^{n-1} \sum_{h=1}^{d} \frac{h{d \choose h} p^{d-h} (1-p)^{h}}{ \gamma_{r,h} }, %
        \\
            \gamma_{r,h}  & \triangleq  & 4 (1 +r \cdot h) \cdot \left(\sqrt{1 + r \cdot h + h} 
 - \sqrt{1 + r \cdot h}\right)^2,\\
         \tilde{h} &\triangleq &\left\lceil \nicefrac{h_{\max}(1-p)}{n}+ \sqrt{\nicefrac{3h_{\max}(1-p)}{n}   \log \left( {\nicefrac{1}{\delta'}} \right)} \right\rceil,
    \\
    \alpha& \triangleq &\Scale[0.99]{\min\left(\frac{\sqrt{2\log (\nicefrac{1}{\delta}) \log (\nicefrac{1.25}{\delta})}}{\varepsilon\sqrt{a}}+1,\frac{1+\sqrt{\frac{16  \log \left(\nicefrac{1.25}{\delta}\right)}{\varepsilon^2}+1}}{2}\right)},\IEEEeqnarraynumspace
  \end{IEEEeqnarray*}
  and $0 \leq p < 1$ is the probability of skipping a node.\footnote{For the uniform random walk scheme considered in \cite{CyffersBellet22_1}, a similar result can be derived (see \cref{thm:ss_rand_walk_all_noise} in Appendix~\ref{sec:uniform_random_walk} in the supplementary material). In fact, the only distinction lies in a different definition of the parameter $a$. However, as shown there, the privacy leakage level $\varepsilon_{\mathrm{skip}}$ is higher compared to the \ssrandring\ scheme.}
\end{theorem}

\begin{remark} \label{rem:1}
It follows from \cref{thm:ss_ring_all_noise,thm:ss_rand_ring_all_noise} that the asymptotic behavior of the privacy leakage level $\varepsilon_{\mathrm{skip}}$ for both \ssring\ and \ssrandring\ is linear in $h_{\max}$, i.e., $\varepsilon_{\mathrm{skip}} = O(h_{\max})$, for $0 \leq p < 1$.
\end{remark}

As a final  remark, the privacy analysis relies on the exact number of updates performed. Skipping introduces uncertainty on which nodes participated and can be seen as a way to realize \emph{subsampling} \cite{BalleBartheGaboardi18_1} on the fly. %

\section{Experiments}
\label{sec:simulations}

Here, we first present some results on the computation and communication latency for the skipping scheme in \cref{alg:ss} that will be used in the numerical results.

Second, we  perform a comparison based on  the analytical results from \cref{sec:convergence-analysis,sec:privacy}, before turning to training a logistic regression model using the dataset in~\cite{OpenMLdata14_1} and a deep neural network for image classification using the MNIST~\cite{LecunBottouBengioHaffner98_1} and CIFAR-$10$~\cite{Krizhevsky09_1} datasets. Finally, we compare with  a parallel and a centralized federated learning approach.

\subsection{Computation and Communication Latency}
\label{sec:comp-comm-latency}

The average total latency of the skipping scheme in \cref{alg:ss} is given by the following lemma.

\begin{lemma} \label{prop:latency}
  The expected total latency for the skipping scheme in \cref{alg:ss} is
  \begin{IEEEeqnarray*}{c}
    h_{\max} \left(\chi + \int_0^{\tss} t \,\mathrm{d}\Phi_T(t) %
      + \tss \bigl(1-\Phi_T(\tss)\bigr)\right),
  \end{IEEEeqnarray*}
  where $\Phi_T(t)\eqdef\Pr[T \leq t]$ and $\Phi_T(\tss)=1-p$.
\end{lemma}

If the number of  hops $h_{\max}$ is large enough, 
we would expect shorter times between token updates (all other properties being the same) to be beneficial for convergence. In other words,
expected time between two consecutive visits to \cref{alg:ss:token-update} in \cref{alg:ss} should be minimized.%

\begin{lemma}
  \label{prop:optimal_tss}
	The value of $\tss$ that minimizes the average time between two consecutive updates of the token is given by the solution of the optimization problem\footnote{Note that the \emph{optimal} value of $\tss$  can incorporate the probability of link failures and channel noise between nodes by changing the distribution of $T$.}
\begin{IEEEeqnarray*}{c} %
	\argmin_{\tss} \frac{\chi + \int_0^{\tss} t \,\mathrm{d}\Phi_T(t) %
		+ \tss \bigl(1-\Phi_T(\tss)\bigr)}{\Phi_T(\tss)}.
	\end{IEEEeqnarray*}
\end{lemma}

\subsection{Convergence Versus Privacy and Average Latency}
\label{sec:numerical_results_theory}

\begin{figure*}[t]
	\centering
	\subfloat{
\begin{tikzpicture}

\definecolor{darkgray176}{RGB}{176,176,176}
\definecolor{lightgray204}{RGB}{204,204,204}

\pgfplotsset{set layers}

\begin{axis}[
	scale only axis,
	axis y line*=right,
	legend cell align={left},
legend style={fill opacity=0.8, draw opacity=1, text opacity=1, draw=lightgray204, font=\tiny, at={(0.53,0.99)}},
	grid style={gray,opacity=0.5,dotted},
	xmajorgrids,
	ymax=16,
	ymin=0,
	xmin=0,
	xmax=1000,
	width=\mywidth,
	height=\myheight,
	tick label style={font=\small},
]

\addplot [semithick,red] table [mark=none,col sep=comma] {figs/curves/EPSvsLAT_Expon_n_10_deltap_0.000001_ring_p_0.0001.csv};
\addlegendentry{$p=10^{-4}$}

\addplot [semithick,blue] table [mark=none,col sep=comma] {figs/curves/EPSvsLAT_Expon_n_10_deltap_0.000001_ring_p_0.5000.csv};
\addlegendentry{$p=\nicefrac{1}{2}$}

\addplot [semithick,black] table [mark=none,col sep=comma] {figs/curves/EPSvsLAT_Expon_n_10_deltap_0.000001_ring_p_0.7000.csv};
\addlegendentry{$p=\nicefrac{7}{10}$}

\addplot [semithick,red,dashed] table [mark=none,col sep=comma] {figs/curves/EPSvsLAT_Expon_n_10_deltap_0.000001_randring_p_0.0001.csv};
\addplot [semithick,blue,dashed] table [mark=none,col sep=comma] {figs/curves/EPSvsLAT_Expon_n_10_deltap_0.000001_randring_p_0.5000.csv};
\addplot [semithick,black,dashed] table [mark=none,col sep=comma] {figs/curves/EPSvsLAT_Expon_n_10_deltap_0.000001_randring_p_0.7000.csv};

\end{axis}

\begin{axis}[
	scale only axis,
	axis y line*=left,
	axis x line=none,
	ymax=1500,
	ymin=0,
	scaled y ticks=base 10:-3,
	xmin=0,
	xmax=1000,
	ylabel={Expected error},
	width=\mywidth,
	height=\myheight,
	tick label style={font=\small},
]

\addplot [semithick,red] table [mark=none,col sep=comma] {figs/curves/ERRvsLAT_Expon_n_10_deltap_0.000001_ring_p_0.0001.csv};
\addplot [semithick,blue] table [mark=none,col sep=comma] {figs/curves/ERRvsLAT_Expon_n_10_deltap_0.000001_ring_p_0.5000.csv};
\addplot [semithick,black] table [mark=none,col sep=comma] {figs/curves/ERRvsLAT_Expon_n_10_deltap_0.000001_ring_p_0.7000.csv};

\addplot [semithick,red,dashed] table [mark=none,col sep=comma] {figs/curves/ERRvsLAT_Expon_n_10_deltap_0.000001_randring_p_0.0001.csv};
\addplot [semithick,blue,dashed] table [mark=none,col sep=comma] {figs/curves/ERRvsLAT_Expon_n_10_deltap_0.000001_randring_p_0.5000.csv};
\addplot [semithick,black,dashed] table [mark=none,col sep=comma] {figs/curves/ERRvsLAT_Expon_n_10_deltap_0.000001_randring_p_0.7000.csv};

\end{axis}

\end{tikzpicture}
	}
	\hfill
	\subfloat{
\begin{tikzpicture}
	
	\definecolor{darkgray176}{RGB}{176,176,176}
	\definecolor{lightgray204}{RGB}{204,204,204}
	
	\pgfplotsset{set layers}

	\begin{axis}[
		scale only axis,
		axis y line*=right,
		legend cell align={left},
		legend style={fill opacity=0.8, draw opacity=1, text opacity=1, draw=lightgray204, font=\tiny, at={(0.53,0.99)}},
		grid style={gray,opacity=0.5,dotted},
		xmajorgrids,
		ymax=80,
		ymin=0,
		xmin=0,
		xmax=1000,
		width=\mywidth,
		height=\myheight,
		tick label style={font=\small},
		]
		
		\addplot [semithick,red] table [mark=none,col sep=comma] {figs/curves/EPSvsLAT_Gamma_0.25_1_n_10_deltap_0.000001_ring_p_0.0001.csv};
		\addlegendentry{$p=10^{-4}$}
		
		\addplot [semithick,blue] table [mark=none,col sep=comma] {figs/curves/EPSvsLAT_Gamma_0.25_1_n_10_deltap_0.000001_ring_p_0.5000.csv};
		\addlegendentry{$p=\nicefrac{1}{2}$}
		
		\addplot [semithick,black] table [mark=none,col sep=comma] {figs/curves/EPSvsLAT_Gamma_0.25_1_n_10_deltap_0.000001_ring_p_0.7000.csv};
		\addlegendentry{$p=\nicefrac{7}{10}$}
		
		\addplot [semithick,red,dashed] table [mark=none,col sep=comma] {figs/curves/EPSvsLAT_Gamma_0.25_1_n_10_deltap_0.000001_randring_p_0.0001.csv};
		\addplot [semithick,blue,dashed] table [mark=none,col sep=comma] {figs/curves/EPSvsLAT_Gamma_0.25_1_n_10_deltap_0.000001_randring_p_0.5000.csv};
		\addplot [semithick,black,dashed] table [mark=none,col sep=comma] {figs/curves/EPSvsLAT_Gamma_0.25_1_n_10_deltap_0.000001_randring_p_0.7000.csv};
		
	\end{axis}
	
	\begin{axis}[
		scale only axis,
		axis y line*=left,
		axis x line=none,
		ymax=1500,
		ymin=0,
		scaled y ticks=base 10:-3,
		xmin=0,
		xmax=1000,
		width=\mywidth,
		height=\myheight,
		tick label style={font=\small},
		]
		
		\addplot [semithick,red] table [mark=none,col sep=comma] {figs/curves/ERRvsLAT_Gamma_0.25_1_n_10_deltap_0.000001_ring_p_0.0001.csv};
		\addplot [semithick,blue] table [mark=none,col sep=comma] {figs/curves/ERRvsLAT_Gamma_0.25_1_n_10_deltap_0.000001_ring_p_0.5000.csv};
		\addplot [semithick,black] table [mark=none,col sep=comma] {figs/curves/ERRvsLAT_Gamma_0.25_1_n_10_deltap_0.000001_ring_p_0.7000.csv};
		
		\addplot [semithick,red,dashed] table [mark=none,col sep=comma] {figs/curves/ERRvsLAT_Gamma_0.25_1_n_10_deltap_0.000001_randring_p_0.0001.csv};
		\addplot [semithick,blue,dashed] table [mark=none,col sep=comma] {figs/curves/ERRvsLAT_Gamma_0.25_1_n_10_deltap_0.000001_randring_p_0.5000.csv};
		\addplot [semithick,black,dashed] table [mark=none,col sep=comma] {figs/curves/ERRvsLAT_Gamma_0.25_1_n_10_deltap_0.000001_randring_p_0.7000.csv};
		
	\end{axis}
	
\end{tikzpicture}
	}
	\hfill
	\subfloat{
\begin{tikzpicture}
	
	\definecolor{darkgray176}{RGB}{176,176,176}
	\definecolor{lightgray204}{RGB}{204,204,204}
	
	\pgfplotsset{set layers}

	\begin{axis}[
		scale only axis,
		axis y line*=right,
		legend cell align={left},
		legend style={fill opacity=0.8, draw opacity=1, text opacity=1, draw=lightgray204, font=\tiny, at={(0.53,0.99)}},
		grid style={gray,opacity=0.5,dotted},
		xmajorgrids,
		ymax=16,
		ymin=0,
		xmin=0,
		xmax=1000,
		ylabel={$\varepsilon_{\mathrm{skip}}$},
		width=\mywidth,
		height=\myheight,
		tick label style={font=\small},
		]
		
		\addplot [semithick,red] table [mark=none,col sep=comma] {figs/curves/EPSvsLAT_Pareto_3_2_n_10_deltap_0.000001_ring_p_0.0001.csv};
		\addlegendentry{$p=10^{-4}$}
		
		\addplot [semithick,blue] table [mark=none,col sep=comma] {figs/curves/EPSvsLAT_Pareto_3_2_n_10_deltap_0.000001_ring_p_0.5000.csv};
		\addlegendentry{$p=\nicefrac{1}{2}$}
		
		\addplot [semithick,black] table [mark=none,col sep=comma] {figs/curves/EPSvsLAT_Pareto_3_2_n_10_deltap_0.000001_ring_p_0.7000.csv};
		\addlegendentry{$p=\nicefrac{7}{10}$}
		
		\addplot [semithick,red,dashed] table [mark=none,col sep=comma] {figs/curves/EPSvsLAT_Pareto_3_2_n_10_deltap_0.000001_randring_p_0.0001.csv};
		\addplot [semithick,blue,dashed] table [mark=none,col sep=comma] {figs/curves/EPSvsLAT_Pareto_3_2_n_10_deltap_0.000001_randring_p_0.5000.csv};
		\addplot [semithick,black,dashed] table [mark=none,col sep=comma] {figs/curves/EPSvsLAT_Pareto_3_2_n_10_deltap_0.000001_randring_p_0.7000.csv};
		
	\end{axis}
	
	\begin{axis}[
		scale only axis,
		axis y line*=left,
		axis x line=none,
		ymax=1500,
		ymin=0,
		scaled y ticks=base 10:-3,
		xmin=0,
		xmax=1000,
		width=\mywidth,
		height=\myheight,
		tick label style={font=\small},
		]
		
		\addplot [semithick,red] table [mark=none,col sep=comma] {figs/curves/ERRvsLAT_Pareto_3_2_n_10_deltap_0.000001_ring_p_0.0001.csv};
		\addplot [semithick,blue] table [mark=none,col sep=comma] {figs/curves/ERRvsLAT_Pareto_3_2_n_10_deltap_0.000001_ring_p_0.5000.csv};
		\addplot [semithick,black] table [mark=none,col sep=comma] {figs/curves/ERRvsLAT_Pareto_3_2_n_10_deltap_0.000001_ring_p_0.7000.csv};
		
		\addplot [semithick,red,dashed] table [mark=none,col sep=comma] {figs/curves/ERRvsLAT_Pareto_3_2_n_10_deltap_0.000001_randring_p_0.0001.csv};
		\addplot [semithick,blue,dashed] table [mark=none,col sep=comma] {figs/curves/ERRvsLAT_Pareto_3_2_n_10_deltap_0.000001_randring_p_0.5000.csv};
		\addplot [semithick,black,dashed] table [mark=none,col sep=comma] {figs/curves/ERRvsLAT_Pareto_3_2_n_10_deltap_0.000001_randring_p_0.7000.csv};
		
	\end{axis}
	
\end{tikzpicture}
	}
	\\
	\vspace{-4.0mm}
	\addtocounter{subfigure}{-3}
	\hspace{1.5mm}
	\subfloat[Exponential (mean $1$).]{
\begin{tikzpicture}

\definecolor{darkgray176}{RGB}{176,176,176}
\definecolor{lightgray204}{RGB}{204,204,204}

\pgfplotsset{set layers}

\begin{axis}[
	scale only axis,
	axis y line*=right,
	legend cell align={left},
	legend style={fill opacity=0.8, draw opacity=1, text opacity=1, draw=lightgray204, font=\tiny, at={(0.53,0.99)}},
	grid style={gray,opacity=0.5,dotted},
	xmajorgrids,
	ymax=8,
	ymin=0,
	xmin=0,
	xmax=6000,
	xlabel={Average latency},
	width=\mywidth,
	height=\myheight,
	tick label style={font=\small},
]

\addplot [semithick,red] table [mark=none,col sep=comma] {figs/curves/EPSvsLAT_Expon_n_500_deltap_0.000001_ring_p_0.0001.csv};
\addlegendentry{$p=10^{-4}$}

\addplot [semithick,blue] table [mark=none,col sep=comma] {figs/curves/EPSvsLAT_Expon_n_500_deltap_0.000001_ring_p_0.5000.csv};
\addlegendentry{$p=\nicefrac{1}{2}$}

\addplot [semithick,black] table [mark=none,col sep=comma] {figs/curves/EPSvsLAT_Expon_n_500_deltap_0.000001_ring_p_0.7000.csv};
\addlegendentry{$p=\nicefrac{7}{10}$}

\addplot [semithick,red,dashed] table [mark=none,col sep=comma] {figs/curves/EPSvsLAT_Expon_n_500_deltap_0.000001_randring_p_0.0001.csv};
\addplot [semithick,blue,dashed] table [mark=none,col sep=comma] {figs/curves/EPSvsLAT_Expon_n_500_deltap_0.000001_randring_p_0.5000.csv};
\addplot [semithick,black,dashed] table [mark=none,col sep=comma] {figs/curves/EPSvsLAT_Expon_n_500_deltap_0.000001_randring_p_0.7000.csv};

\end{axis}

\begin{axis}[
	scale only axis,
	axis y line*=left,
	axis x line=none,
	ymax=4000,
	ymin=0,
	scaled y ticks=base 10:-3,
	xmin=0,
	xmax=6000,
	xtick style={color=black},
	ylabel={Expected error},
	width=\mywidth,
	height=\myheight,
	tick label style={font=\small},
]

\addplot [semithick,red] table [mark=none,col sep=comma] {figs/curves/ERRvsLAT_Expon_n_500_deltap_0.000001_ring_p_0.0001.csv};
\addplot [semithick,blue] table [mark=none,col sep=comma] {figs/curves/ERRvsLAT_Expon_n_500_deltap_0.000001_ring_p_0.5000.csv};
\addplot [semithick,black] table [mark=none,col sep=comma] {figs/curves/ERRvsLAT_Expon_n_500_deltap_0.000001_ring_p_0.7000.csv};

\addplot [semithick,red,dashed] table [mark=none,col sep=comma] {figs/curves/ERRvsLAT_Expon_n_500_deltap_0.000001_randring_p_0.0001.csv};
\addplot [semithick,blue,dashed] table [mark=none,col sep=comma] {figs/curves/ERRvsLAT_Expon_n_500_deltap_0.000001_randring_p_0.5000.csv};
\addplot [semithick,black,dashed] table [mark=none,col sep=comma] {figs/curves/ERRvsLAT_Expon_n_500_deltap_0.000001_randring_p_0.7000.csv};

\end{axis}

\end{tikzpicture}}
	\hfill
	\subfloat[Gamma (shape $\nicefrac 14$, scale $1$).]{
\begin{tikzpicture}
	
	\definecolor{darkgray176}{RGB}{176,176,176}
	\definecolor{lightgray204}{RGB}{204,204,204}
	
	\pgfplotsset{set layers}

	\begin{axis}[
		scale only axis,
		axis y line*=right,
		legend cell align={left},
		legend style={fill opacity=0.8, draw opacity=1, text opacity=1, draw=lightgray204, font=\tiny, at={(0.53,0.99)}},
		grid style={gray,opacity=0.5,dotted},
		xmajorgrids,
		ymin=0,
		xmin=0,
		xmax=1000,
		xlabel={Average latency},
		width=\mywidth,
		height=\myheight,
		tick label style={font=\small},
		]
		
		\addplot [semithick,red] table [mark=none,col sep=comma] {figs/curves/EPSvsLAT_Gamma_0.25_1_n_500_deltap_0.000001_ring_p_0.0001.csv};
		\addlegendentry{$p=10^{-4}$}
		
		\addplot [semithick,blue] table [mark=none,col sep=comma] {figs/curves/EPSvsLAT_Gamma_0.25_1_n_500_deltap_0.000001_ring_p_0.5000.csv};
		\addlegendentry{$p=\nicefrac{1}{2}$}
		
		\addplot [semithick,black] table [mark=none,col sep=comma] {figs/curves/EPSvsLAT_Gamma_0.25_1_n_500_deltap_0.000001_ring_p_0.7000.csv};
		\addlegendentry{$p=\nicefrac{7}{10}$}
		
		\addplot [semithick,red,dashed] table [mark=none,col sep=comma] {figs/curves/EPSvsLAT_Gamma_0.25_1_n_500_deltap_0.000001_randring_p_0.0001.csv};
		\addplot [semithick,blue,dashed] table [mark=none,col sep=comma] {figs/curves/EPSvsLAT_Gamma_0.25_1_n_500_deltap_0.000001_randring_p_0.5000.csv};
		\addplot [semithick,black,dashed] table [mark=none,col sep=comma] {figs/curves/EPSvsLAT_Gamma_0.25_1_n_500_deltap_0.000001_randring_p_0.7000.csv};
		
	\end{axis}
	
	\begin{axis}[
		scale only axis,
		axis y line*=left,
		axis x line=none,
		ymax=4000,
		ymin=0,
		scaled y ticks=base 10:-3,
		xmin=0,
		xmax=1000,
		width=\mywidth,
		height=\myheight,
		tick label style={font=\small},
		]
		
		\addplot [semithick,red] table [mark=none,col sep=comma] {figs/curves/ERRvsLAT_Gamma_0.25_1_n_500_deltap_0.000001_ring_p_0.0001.csv};
		\addplot [semithick,blue] table [mark=none,col sep=comma] {figs/curves/ERRvsLAT_Gamma_0.25_1_n_500_deltap_0.000001_ring_p_0.5000.csv};
		\addplot [semithick,black] table [mark=none,col sep=comma] {figs/curves/ERRvsLAT_Gamma_0.25_1_n_500_deltap_0.000001_ring_p_0.7000.csv};
		
		\addplot [semithick,red,dashed] table [mark=none,col sep=comma] {figs/curves/ERRvsLAT_Gamma_0.25_1_n_500_deltap_0.000001_randring_p_0.0001.csv};
		\addplot [semithick,blue,dashed] table [mark=none,col sep=comma] {figs/curves/ERRvsLAT_Gamma_0.25_1_n_500_deltap_0.000001_randring_p_0.5000.csv};
		\addplot [semithick,black,dashed] table [mark=none,col sep=comma] {figs/curves/ERRvsLAT_Gamma_0.25_1_n_500_deltap_0.000001_randring_p_0.7000.csv};
		
	\end{axis}
	
\end{tikzpicture}}
	\hfill
	\hspace{-1.9mm}
	\subfloat[Pareto type II (shape $3$, scale $2$).]{
\begin{tikzpicture}
	
	\definecolor{darkgray176}{RGB}{176,176,176}
	\definecolor{lightgray204}{RGB}{204,204,204}
	
	\pgfplotsset{set layers}

	\begin{axis}[
		scale only axis,
		axis y line*=right,
		legend cell align={left},
		legend style={fill opacity=0.8, draw opacity=1, text opacity=1, draw=lightgray204, font=\tiny, at={(0.53,0.99)}},
		grid style={gray,opacity=0.5,dotted},
		xmajorgrids,
		ymax=8,
		ymin=0,
		xmin=0,
		xmax=6000,
		xlabel={Average latency},
		ylabel={$\varepsilon_{\mathrm{skip}}$},
		width=\mywidth,
		height=\myheight,
		tick label style={font=\small},
		]
		
		\addplot [semithick,red] table [mark=none,col sep=comma] {figs/curves/EPSvsLAT_Pareto_3_2_n_500_deltap_0.000001_ring_p_0.0001.csv};
		\addlegendentry{$p=10^{-4}$}
		
		\addplot [semithick,blue] table [mark=none,col sep=comma] {figs/curves/EPSvsLAT_Pareto_3_2_n_500_deltap_0.000001_ring_p_0.5000.csv};
		\addlegendentry{$p=\nicefrac{1}{2}$}
		
		\addplot [semithick,black] table [mark=none,col sep=comma] {figs/curves/EPSvsLAT_Pareto_3_2_n_500_deltap_0.000001_ring_p_0.7000.csv};
		\addlegendentry{$p=\nicefrac{7}{10}$}
		
		\addplot [semithick,red,dashed] table [mark=none,col sep=comma] {figs/curves/EPSvsLAT_Pareto_3_2_n_500_deltap_0.000001_randring_p_0.0001.csv};
		\addplot [semithick,blue,dashed] table [mark=none,col sep=comma] {figs/curves/EPSvsLAT_Pareto_3_2_n_500_deltap_0.000001_randring_p_0.5000.csv};
		\addplot [semithick,black,dashed] table [mark=none,col sep=comma] {figs/curves/EPSvsLAT_Pareto_3_2_n_500_deltap_0.000001_randring_p_0.7000.csv};
		
	\end{axis}
	
	\begin{axis}[
		scale only axis,
		axis y line*=left,
		axis x line=none,
		ymax=4000,
		ymin=0,
		scaled y ticks=base 10:-3,
		xmin=0,
		xmax=6000,
		width=\mywidth,
		height=\myheight,
		tick label style={font=\small},
		]
		
		\addplot [semithick,red] table [mark=none,col sep=comma] {figs/curves/ERRvsLAT_Pareto_3_2_n_500_deltap_0.000001_ring_p_0.0001.csv};
		\addplot [semithick,blue] table [mark=none,col sep=comma] {figs/curves/ERRvsLAT_Pareto_3_2_n_500_deltap_0.000001_ring_p_0.5000.csv};
		\addplot [semithick,black] table [mark=none,col sep=comma] {figs/curves/ERRvsLAT_Pareto_3_2_n_500_deltap_0.000001_ring_p_0.7000.csv};
		
		\addplot [semithick,red,dashed] table [mark=none,col sep=comma] {figs/curves/ERRvsLAT_Pareto_3_2_n_500_deltap_0.000001_randring_p_0.0001.csv};
		\addplot [semithick,blue,dashed] table [mark=none,col sep=comma] {figs/curves/ERRvsLAT_Pareto_3_2_n_500_deltap_0.000001_randring_p_0.5000.csv};
		\addplot [semithick,black,dashed] table [mark=none,col sep=comma] {figs/curves/ERRvsLAT_Pareto_3_2_n_500_deltap_0.000001_randring_p_0.7000.csv};
		
	\end{axis}
	
\end{tikzpicture}}
	\hspace{1.5mm}
	\vspace{-0ex}
	\caption{Expected error bound (decreasing curves; \cref{thm:convergence}) and privacy leakage level $\varepsilon_{\mathrm{skip}}$ (increasing curves; \cref{thm:ss_ring_all_noise,thm:ss_rand_ring_all_noise}) vs average latency (\cref{prop:latency}) for $n=10$ (top row) and $n=500$ (bottom row). Solid lines are for a fixed ring (\ssring), while dashed lines are for \ssrandring.
	}
	\label{fig:EPSandERRvsLAT}
	\vspace{-0mm}
\end{figure*}

\begin{figure*}[t]
	\centering
	\includegraphics[width=1.0\textwidth]{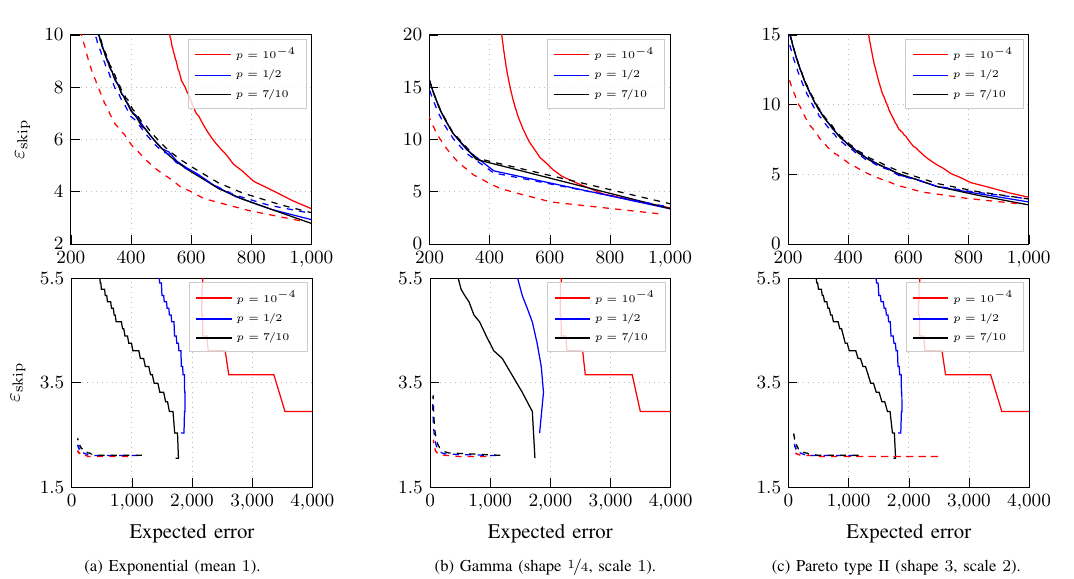}
	\vspace{-4ex}
	\caption{Privacy leakage level $\varepsilon_{\mathrm{skip}}$ vs expected error bound for $n=10$ (top row) and $n=500$ (bottom row). Solid lines are for a fixed ring (\ssring), while dashed lines are for \ssrandring.}
	\label{fig:EPSvsERR}
	\vspace{-0mm}
\end{figure*}

We fix  $\varepsilon=1$, $\delta=10^{-6}$, $\delta'=10^{-6}$, $d=8$, $d_\set{W}=10$, $k=1$, 
$\zeta=\nicefrac{6}{10}$, 
and $\chi=\nicefrac{1}{100}$. Results are presented for two different values of the number of nodes $n$, namely for a small number of $n=10$ nodes and a large number of  $n=500$  nodes.\footnote{In \cite{KoloskovaLoizouBoreiriJaggiStich20_1} and \cite{ChenDahlLarsson23_1, HerreraChenLarsson24_1sub}, a rather small number of nodes ($n=25$ and $n=15$ or $20$, respectively) was used in all numerical results, while in \cite{CyffersBellet22_1, CyffersEvenBelletMassoulie22_1} a rather large number of nodes ($n=1000$, $2000$, or $4000$) was used.}
The three characteristics we are interested in are: average latency, expected error bound, and privacy leakage level $\varepsilon_{\mathrm{skip}}$. We first consider $n=10$ nodes. The top row of \cref{fig:EPSandERRvsLAT} %
plots expected error bound (left $y$-axis) and privacy leakage level (right $y$-axis) versus average latency, and the top row of \cref{fig:EPSvsERR} shows privacy leakage level versus expected error bound, illustrating the inherent trade-off between average latency, expected error bound, and privacy leakage level. The plots are for the three  latency models: exponential with mean $1$, gamma with shape $\nicefrac 14$ and scale $1$, and Pareto type II with shape $3$ and scale $2$ (as used in \cite{NegliaCalbiTowsleyVardoyan19_1}). %
The probability of skipping  $p = \Pr[T > \tss] \in \{10^{-4},\nicefrac{1}{2}, \nicefrac{7}{10} \}$,  since $p=10^{-4}$ and $\nicefrac{7}{10}$ are close to the values of $p$ corresponding to the optimal values of $\tss$ given by \cref{prop:optimal_tss}, respectively $0/0.710/0.737$ for the exponential/gamma/Pareto delay models, while $p=\nicefrac{1}{2}$ is a value in between.\footnote{We have picked $p=10^{-4}$ instead of $p=0$ as \cref{thm:convergence} requires $p>0$  in the \ssring\ scheme.}

As can be seen from the plots, $p=10^{-4}$ (virtually, no skipping) gives the worst expected error bound for all considered latency models, for both schemes. This is particularly evident for \ssring{} with $n=500$ (second row of plots), where the convergence rate is noticeably slow due to $|\lambda_1| \approx 1 - 10^{-8}$, which is very close to $1$. On the other hand, this value of $p$ provides 
the best privacy leakage level for the same average latency. Hence, there is a trade-off between privacy and accuracy of the algorithm (cf.~the top row of plots in \cref{fig:EPSvsERR}), and one needs to choose the skipping probability based on a particular optimization problem.

The privacy-versus-error trade-offs look similar for all latency models considered. \ssrandring\ gives better trade-off curves (especially for $p=10^{-4}$) for smaller values of expected error bound, while the situation changes for higher values of error (i.e., at the initial stages of \cref{alg:ss}'s execution).	
	Hence, path randomization improves the trade-off in the long run,  but might be harder to realize in a real-world implementation as it would require a full mesh topology.\footnote{Strictly speaking a full mesh topology is also required for \ssring, as for a high skipping probability $p$ there could potentially be a need for every single node to be able to communicate with all other nodes, while  with no skipping only one output communication channel per node is required. However, as $p$ is constant, and the unavailability is assumed independent from one node to another, a few edges should guarantee that at least one node will answer. The probability that more than $l$ edges would be required is $p^l$, which quickly becomes small, e.g., for $p=\nicefrac{1}{2}$ and $10$ edges, the probability is less than $10^{-3}$.}
	
{On the contrary, the \ssring\ curve for $p=10^{-4}$ is the worst, which means that skipping helps. Also, there is not much difference between the \ssring\ curves for $p=\nicefrac 12$ and $p=\nicefrac{7}{10}$ (they are are almost on top of each other and hence difficult to distinguish). 
On the other hand, \ssrandring\ favors smaller values of $p$ (i.e., larger timeout) at the expense of a higher training latency as shown in the next subsection.

In the bottom rows of \cref{fig:EPSandERRvsLAT,fig:EPSvsERR}, we show the corresponding results for $n=500$ nodes. As expected, the relative order of the curves remains for the most part the same as for $n=10$ nodes (compare with the top rows of the figures). We also observe from \cref{fig:EPSvsERR} that for a given expected error bound the privacy leakage level $\varepsilon_{\mathrm{skip}}$ is lower with $n=500$ than with $n=10$ nodes, i.e., privacy amplification kicks in to a larger extent with a larger number of nodes. Also, the \ssrandring\ scheme shows in general a much bigger privacy advantage compared to the \ssring\ scheme as the privacy amplification effect is stronger with randomization. Finally, note the more pronounced staircase behavior for the privacy leakage level. This is due to the factor $\nicefrac{(1-p)}{n}$ inside the ceiling function in the definition of $\tilde{h}$ in \cref{thm:ss_ring_all_noise,thm:ss_rand_ring_all_noise}, %
which also explains why the steps are wider for %
a larger $n$.

\subsection{Empirical Results} \label{sec:numerical_results_regression}

We consider both  training a logistic regression model and image classification trained on the MNIST~\cite{LecunBottouBengioHaffner98_1} and CIFAR-$10$~\cite{Krizhevsky09_1} datasets.

\subsubsection{Logistic Regression} \label{sec:numerical_logistic_regression}

For logistic regression the local loss functions are   $f_v(\tau,\set D_v) = \nicefrac{1}{|\set D_v|} \sum_{(x,y) \in \set D_v} \log(1+{\mathrm{e}}^{-y \tau x^\top})$, where $x \in \Reals^{d}$ ($d_\mathrm{x}=d$) and $y \in \{-1,1\}$ ($d_\mathrm{y}=1$).
We use a binarized version of the UCI housing dataset~\cite{OpenMLdata14_1}, trying to predict binary variable $y$ (whether house price is above a threshold) from other features, $x$. The features %
are standardized and we further normalize each data point to have unit $\ell_2$-norm so that the  loss functions $f_v(\tau;\set D_v)$ are $1$-Lipschitz continuous (i.e., $k=1$). The dataset is split uniformly at random into a training set with $80\%$ of the data points  and a test set with $20\%$ of the points. Moreover, the training dataset is further randomly split  across the $n$ nodes in $\set V$. We used the \ssrandring\ scheme (similar results are obtained with the \ssring\ scheme) with the same parameters as in \cref{sec:numerical_results_theory}, but 
using a mini-batch implementation with batches of size $100$ and $8$ and with an initial learning rate of 
$\zeta=\nicefrac{6}{10}$ and $\zeta=\nicefrac{3}{10}$  
for, respectively, $n=10$ and $n=1000$ nodes in order to speed up the learning. 
The chosen mini-batch size is a compromise between the two corner cases: 
a mini-batch size of $1$ is difficult to parallelize, whereas a large mini-batch size may exceed the nodes' limited parallelization capabilities.

\begin{figure*}[t]
  \centering
  \subfloat{%
    \input{MarkFigures/jsaitNewFig/expSkipRingn10bn100errBarRun200.tex}}
  \subfloat{
    \input{MarkFigures/jsaitNewFig/gammaSkipRingn10bn100errBarRun200.tex}}
  \hspace{0.01cm}
  \subfloat{          
    \input{MarkFigures/jsaitNewFig/paretoSkipRingn10bn100errBarRun200.tex}}
  \\
  \vspace{-5mm}
       \subfloat{%
       \input{MarkFigures/jsaitNewFig/expSkipRingn1000bn8errBarRun200.tex}}
  \subfloat{
    \input{MarkFigures/jsaitNewFig/gammaSkipRingn1000bn8errBarRun200.tex}}
  \hspace{0.01cm}
  \subfloat{          
    \input{MarkFigures/jsaitNewFig/paretoSkipRingn1000bn8errBarRun200.tex}}
  \\
  \vspace{-5mm}
  \subfloat{
    \input{MarkFigures/MNISTexpR30Samp10errbar.tex}}%
  \subfloat{
    \input{MarkFigures/MNISTgammaR30Samp10errbar.tex}
  }%
  \subfloat{          
    \input{MarkFigures/MNISTparetoR30Samp10errbar.tex}\hspace{0.40cm}
  }%
  \\
  \vspace{-5mm}
  \setcounter{subfigure}{0}
  \subfloat[Exponential (mean $1$).]{\hspace{0.02cm}
    \input{MarkFigures/CIFARexpR6Samp10errbar.tex}\hspace{0.14cm}}
  \subfloat[Gamma (shape $\nicefrac 14$, scale $1$).]{%
    \input{MarkFigures/CIFARgammaR6Samp10errbar.tex}}
  \subfloat[Pareto type II (shape $3$, scale $2$).]{\hspace{0.10cm}          
    \input{MarkFigures/CIFARparetoR6Samp10errbar.tex}\hspace{0.35cm}
  }%
  \caption{Plots (from top): 1) 
      logistic regression model training with $n=10$ nodes, showing accuracy (on the test set) vs average latency; 2) logistic regression model training with $n=1000$ nodes; 3) image classification using the MNIST dataset  with $n=60$ nodes; 4) image classification using the CIFAR-$10$ dataset with $n=50$ nodes. First and second row of plots: each curve is an average of $200$ independent runs for \ssrandring, while for the third and fourth  row of plots an average of,  respectively, $30$ and $6$ runs is presented (\ssrandring). Horizontal and vertical error bars illustrate the estimated standard deviation. %
  }
  \label{fig:Accuracy_empirical}
\end{figure*} %

\begin{figure*}[t]
  \centering
  \subfloat{\hspace{0.63cm}
    \input{MarkFigures/Fig5a_10.tex}} %
  \subfloat{
    \input{MarkFigures/Fig5b_10.tex}}%
  \vspace{-2ex}
   \caption{Comparing \ssrandring\ (solid curves) with \texttt{Muffliato-SGD}\ (\cite[Alg.~3]{CyffersEvenBelletMassoulie22_1}; dashed and dashdotted curves) and 
   		\texttt{FedL-SGD} (dotted curves) for logistic regression model training on the UCI housing dataset with $n=1000$ nodes for the exponential delay model with mean $1$. Left: test set accuracy vs average latency. Right: privacy leakage level vs average latency.  Each simulation curve for \texttt{Muffliato-SGD}\ and \texttt{FedL-SGD}\ is an average of $100$ independent runs, while for \ssrandring\  the average of $200$ independent runs is presented. In order to not clutter the plots, no error bars are included.}
  \label{fig:Muffliato_comparisons}
\end{figure*}

For $n=10$ nodes, the results of the training are shown in the top plots in \cref{fig:Accuracy_empirical}, which show the prediction error rate, i.e., the ratio of incorrect predictions on the test set,
versus average latency from \cref{prop:latency} for the same skipping probabilities  as in the corresponding plots in \cref{fig:EPSandERRvsLAT,fig:EPSvsERR}. We observe that skipping achieves a clear speed-up compared to no skipping, except for the exponential delay model (as predicted well by \cref{prop:optimal_tss}, which suggests an optimal $\tss=+\infty$ for the exponential model). This rhymes well with theoretical expected error bounds (dashed curves of the plots in \cref{fig:EPSandERRvsLAT}). 
As can be seen from the plots of \cref{fig:EPSvsERR}, no skipping in general provides  a slightly higher privacy for \ssrandring. In the second row of plots in \cref{fig:Accuracy_empirical}, we show the corresponding results with $n=1000$ nodes. As expected, the main conclusions remain the same as for $n=10$. In order to have smooth curves the average of  $200$ independent runs is presented for both $n=10$ and $n=1000$ nodes. %

\subsubsection{Image Classification}

We consider both the MNIST and CIFAR-$10$ datasets. Both datasets are commonly-used benchmarks and are comprised of $10$ classes of images; MNIST being comprised of $28 \times 28$ pixels grayscale images of handwritten  digits from $0$ to $9$, while CIFAR-$10$ being comprised of  $32 \times 32$ pixels color images. The number of training samples is $60000$ ($6000$ for each digit) and $50000$ ($5000$ for each class) for the MNIST and  CIFAR-$10$ datasets, respectively.  As for logistic regression in \cref{sec:numerical_logistic_regression}, the training dataset is further randomly split  across  a number of nodes $n$ in $\set V$. While we used $n=10$ and $n=1000$ nodes  in \cref{sec:numerical_logistic_regression}, we use $n=60$ and $n=50$ nodes, respectively, for the MNIST and CIFAR-$10$ datasets. As for logistic regression, we use the \ssrandring\ scheme with the same parameters as in \cref{sec:numerical_results_theory}, but with a smaller initial learning rate of  $\zeta=\nicefrac{3}{1000}$ (MNIST) and $\zeta=\nicefrac{7}{10000}$ (CIFAR-$10$), and a batch size of $500$, which is half the number of data samples in each node. Moreover, we use a cross-entropy loss function. %

The results are depicted in the third and fourth row of plots in \cref{fig:Accuracy_empirical}, showing the prediction error rate on the test set (comprising $10000$ images for both datasets) versus average latency from \cref{prop:latency}.
For both MNIST (the third row of plots) and CIFAR-$10$ (the bottom plots), we can make the same observations as for the  first and second row of plots (logistic regression); skipping achieves a speed-up compared to no skipping, except for the exponential delay model, as predicted by \cref{prop:optimal_tss}. Moreover, the order of the curves stays the same across the datasets  for a given computational delay model. Note, however, that there is some loss in accuracy due to privacy; the accuracy achieved with the MNIST dataset is close to $90\%$, while with no privacy requirement an accuracy of around $99\%$ can be reached. For the CIFAR-$10$ dataset, the accuracy decreases from around $70\%$ to around $42\%$ in the best case. This aligns well with results in the literature, showing a reduction in accuracy due to privacy, which is particularly significant  for CIFAR-$10$, see, e.g., \cite{de2022unlocking}. 
 Compared to the case of  logistic regression, the average of  only $30$ (MNIST) and $6$ (CIFAR-$10$) independent runs is presented due to the much more complex learning task. The corresponding %
deep neural networks are detailed in Table~\ref{tab:table_NNarch} in the supplementary material.

\subsection{Comparisons With a Parallel and a Centralized Federated Learning Approach}

For completeness, we also compare our results for logistic regression with  a parallel approach using gossip averaging between each step of gradient descent for every node. The most relevant work to compare with is  \cite{CyffersEvenBelletMassoulie22_1}. In \cite[Fig.1(c)]{CyffersEvenBelletMassoulie22_1}, results are presented  for logistic regression on the UCI housing dataset~\cite{OpenMLdata14_1} of \cref{sec:numerical_logistic_regression} using  \cite[Alg.~3]{CyffersEvenBelletMassoulie22_1}  (\texttt{Muffliato-SGD}). We have replicated the setup of \cite[Fig.1(c)]{CyffersEvenBelletMassoulie22_1} (using random Erd\H{o}s-R{\'e}nyi communication graphs with node degree $\log n$ during gossiping), but with $n=1000$ nodes, a fixed number of $2$ gossip iterations, and 
$\gamma=1$ in \cite[Alg.~1]{CyffersEvenBelletMassoulie22_1} (no acceleration)    
and compare  in \cref{fig:Muffliato_comparisons} \texttt{Muffliato-SGD} (dashed and dashdotted  curves) with the \ssrandring\ scheme (solid curves) under the exponential delay model with mean $1$. The left plot shows the error prediction rate on the test set, while the right plot shows the (worst-case) privacy leakage level, both as a function of the average latency  from \cref{prop:latency}. The privacy leakage level for \texttt{Muffliato-SGD} is simulated based on \cite[Thm.~4]{CyffersEvenBelletMassoulie22_1} (for two different values of the privacy noise standard deviation; referred to as instances one and two in the next paragraph) and converting to DP using \cref{lem:RDPtoDP} in Appendix~\ref{app:AppB} with $\delta=10^{-6}$ and with a numerically optimized value of the R{\'e}nyi divergence parameter $\alpha$,
while for \ssrandring\ we have used  the same setup as for the second row of plots in \cref{fig:Accuracy_empirical}, i.e., \cref{thm:ss_rand_ring_all_noise} with  $\epsilon=1.0$, $\delta = 10^{-6}$,  and $\delta' = 10^{-12}$ 
(corresponding to a  DP noise level of $\sigma_h \approx 10.5976$ used in the actual simulation).    
We also compare the prediction error rate and the privacy leakage level with  those of a centralized federated learning approach (dotted curves), denoted by \texttt{FedL-SGD} in the following.\footnote{Our simulation of  \texttt{Muffliato-SGD} and  \texttt{FedL-SGD} is based on \url{https://github.com/totilas/muffliato} where gradient clipping is used. For  \texttt{Muffliato-SGD}, gradient clipping gives  improved accuracy, while for \ssrandring\ we have not observed any noticeable gain with gradient clipping and hence the presented results for \ssrandring\ (as in \cref{fig:Accuracy_empirical}) are with no clipping.}  The (worst-case) privacy leakage level for \texttt{FedL-SGD} is computed as for \texttt{Muffliato-SGD}, by converting to DP using \cref{lem:RDPtoDP} in Appendix~\ref{app:AppB} with $\delta=10^{-6}$ and with a  numerically optimized value of the R{\'e}nyi divergence parameter $\alpha$. In particular, each time a node $u$ uploads to the central server, $\nicefrac{2\alpha}{\sigma_h^2}$ is added to the overall  RDP level of $u$, and the maximum over all nodes $u$ is the worst-case  leakage.
We 
note that implementing gossiping in a latency-efficient manner is not straightforward. In particular, within each iteration of gossiping, each node sends the same information to its neighbors, which can be done through a single broadcast transmission rather than by multiple peer-to-peer transmissions. However, concurrent broadcast transmissions from multiple nodes  create interference, which can lead to failed reception of information at the receiver nodes. A simple solution  would be through a simple time-division approach in which each node broadcasts sequentially. This  entails a communication latency proportional to the number of nodes. A more elaborate approach is  random access with broadcast transmission as outlined in \cite{ChenDahlLarsson23_1} or through broadcast-based subgraph sampling  as outlined in the very recent paper \cite{HerreraChenLarsson24_1sub}. For the results in  \cref{fig:Muffliato_comparisons}, we have used the random access approach outlined in \cite{ChenDahlLarsson23_1} with an optimized value for the probabilistic random access policy. For  \texttt{FedL-SGD}, when computing the training latency, we have assumed $100$ independent subchannels for the upload to the central server and a single broadcast transmission to distribute the aggregated gradient back to the nodes. Having a very large number of subchannels would reduce the bandwidth per channel and hence the transmission rate, assuming a fixed  overall bandwidth constraint \cite{Hu-etal24_1app}, and hence we have used $100$ as a compromise (in \cite{Hu-etal24_1app}, only $8$ or $16$  subchannels were used). More details on the latency computation/simulation %
are given Appendix~\ref{app:additional_num_results} in the supplementary material. 
For \ssrandring, we use $\zeta=\nicefrac{3}{10}$ 
and a batch size of $8$ (as for the second row of plots in \cref{fig:Accuracy_empirical}), while for  \texttt{Muffliato-SGD} and \texttt{FedL-SGD}, we use (as in \cite[Fig.1(c)]{CyffersEvenBelletMassoulie22_1}) a constant learning rate of $\nicefrac{7}{10}$ and a full batch size of $16$ (changing to a batch size of $8$ does not noticeably change the accuracy). For a fair comparison, a  communication cost of $\chi=\nicefrac{1}{100}$ is used for all schemes. %

From \cref{fig:Muffliato_comparisons} (left plot), we observe that the   \ssrandring\ scheme (solid curves) achieves a low error rate quicker than one of the instances of \texttt{Muffliato-SGD} (dashed curves) and  also \texttt{FedL-SGD} with virtually no skipping  (red dotted curves), while for the second instance of \texttt{Muffliato-SGD} with a lower value of the privacy noise standard deviation (dashdotted curves) and for \texttt{FedL-SGD} with skipping ($p=\nicefrac{1}{2}$ and $p=\nicefrac{7}{10}$) we observe the opposite behavior.
On the other hand, the overall privacy leakage level grows much slower with  the   \ssrandring\ scheme (see the right plot). For instance, \texttt{Muffliato-SGD}  (second instance; dashdotted curves) achieves an accuracy of $80\%$ quicker than  \ssrandring\ (in about $6000$ units of time ($p=\nicefrac{1}{2}$) compared to about  $24000$ ($p=10^{-4}$; see first plot in the second row of plots in \cref{fig:Accuracy_empirical}), but at a much higher privacy leakage level ($\varepsilon_{\mathrm{skip}}=5.5$ compared to $2.2$). Compared to the  first instance (dashed curves), however, \ssrandring\ achieves a target accuracy of $80\%$ quicker  but at a lower privacy leakage gap  (the dashed curves in the right plot lie below the dashdotted curves).
\texttt{FedL-SGD} provides a lower  privacy leakage level which also grows slower with latency compared to \texttt{Muffliato-SGD}, but on the other hand  relies on the assumption of a centralized server. The   \ssrandring\ scheme performs favorable compared to \texttt{FedL-SGD} with virtually no skipping, while for $p=\nicefrac{1}{2}$ and $p=\nicefrac{7}{10}$ \texttt{FedL-SGD} yields a lower prediction error rate at the expense of a higher privacy leakage compared to  \ssrandring.  In general,  smaller values of the privacy noise standard deviation $\sigma_h$ for \texttt{FedL-SGD} will provide better accuracy, but at the same time increase the privacy leakage level.

\section{Conclusion and Future Work}

We have studied a skipping scheme for straggler mitigation in decentralized learning over a logical ring under NDP by extending the framework of privacy amplification by decentralization to include overall training latency\textemdash comprising both computation and communication latency. 
Analytical derivations on both the convergence speed and the DP level were presented, showing  a trade-off between  overall training latency, accuracy, and user data privacy. The theoretical findings were validated for logistic regression on a real-world dataset and for image classification using the MNIST and CIFAR-$10$ datasets.

Future work could extend the theoretical analysis in this study to gossip algorithms as examined in \cite{CyffersEvenBelletMassoulie22_1}.

\appendices

\section{Proof of \cref{thm:convergence}}
\label{app:A}

\subsection{Notation}
Define $[a:b]\triangleq \{a,\ldots,b\}$ for integers $a \leq b$. Moreover, $U^*$ denotes the conjugate transpose of a matrix $U$, while $U^{-1}$ denotes its inverse (for a full-rank square matrix $U$). $\diag(a_1,\ldots,a_l)$ denotes an $l \times l$ diagonal matrix with $a_1,\ldots,a_l$ along the diagonal.

\subsection{Preliminaries}

\label{sec:proof-convergence}
For the convergence, what matters is only the nodes that actually contributed to the token updates (nonstragglers, i.e., those that reached \cref{alg:ss:token-update} of \cref{alg:ss}). Let $H \in [0:h_{\max}]$ be the RV denoting the number of nonstragglers when running \cref{alg:ss}, and let the corresponding   nodes visited by the token be denoted by $V^{(1)}, V^{(2)}, \dotsc, V^{(h)}, \dotsc, V^{(H)}$. If $H=0$, then all nodes are straggling, no nodes are visited by the token, and  \cref{alg:ss} simply returns $\tau_0  \triangleq 0$ (i.e., $\tau_{h_{\max}}=\tau_0$). Otherwise (i.e., when $H>0$), according to \cref{alg:ss},
the token updates are (with some abuse of notation)
\[
	\tau_{h} \leftarrow \Pi_{\set W} \left( \tau_{h-1} - \eta_h \left( \nabla f_{V^{(h)}} (\tau_{h-1}; \set D_{V^{(h)}}) + N_h \right) \right),
\]
for all $h \in [H]$. %
Note also that $\eta_h = \nicefrac{\zeta}{\sqrt{h}}$. In the rest of this subsection, we assume $H>0$.

For \ssrandring, the marginal distribution of a node $V^{(h)}$ is uniform
over $\set V$ for any $h$. For \ssring, the sequence of nodes $V^{(1)}, V^{(2)}, \dotsc$ forms a Markov chain with state transition probability matrix
	\begin{equation}
Q = \frac{1-p}{1-p^n}
\begin{pmatrix}
	p^{n-1} & 1       & p      & p^2 & \dots  & p^{n-2} \\
	p^{n-2} & p^{n-1} & 1      & p   & \dots  & p^{n-3} \\
	\vdots  & \vdots  & \vdots &     & \ddots & \vdots \\
	1       & p       & p^2    & p^3 & \dots  & p^{n-1}
\end{pmatrix},\label{eq:Q-matrix}
\end{equation}
where the entries $Q_{ij}\eqdef\Pr[V^{(h)} = v_j \mid V^{(h-1)} = v_i]$, $1\leq i,j\leq n$, $h \geq 1$, and, as we show in \cref{lem:markov-mixing} below, the marginal distributions of $V^{(h)}$ converge to the uniform distribution exponentially fast when $h \to \infty$.

The uniform distribution of $V^{(h)}$ for \ssrandring\ ensures an unbiased estimate of the real (sub)gradient for any fixed $\tau$, i.e.,
\[
\EE[V ^{(h)}]{\nabla f_{V^{(h)}}(\tau; \mathcal D_{V^{(h)}})} = \nabla f(\tau; \mathcal D), 
\]
while for \ssring\ we have that
 \[
\EE[V ^{(h)}]{\nabla f_{V^{(h)}}(\tau; \mathcal D_{V^{(h)}})} \xrightarrow[h \to \infty]{} \nabla f(\tau; \mathcal D).
\] 

Unbiasness of the (sub)gradient estimate at each step is a known condition used to prove convergence of (conventional) stochastic gradient descent. In this appendix, we will show that having \emph{asymptotically} unbiased estimates is sufficient for the convergence of \cref{alg:ss} too. More precisely, we will adapt a proof from \cite[Thm.~2]{ShamirZhang13_1} to our scenario.

First, we present some technical results used in the main part of the proof (next subsection).%

\begin{lemma}\label{lem:markov-mixing}
	For $n \ge 2$, let $\{V^{(h)}\}$, $V^{(h)} \in \set V$, $h \geq 1$, be a homogeneous Markov chain with state transition probability matrix \cref{eq:Q-matrix} with $0 < p < 1$. If we denote by $\pi^{(h)}$ the probability vector of the marginal distribution of $V^{(h)}$ (i.e., $\Pr[V^{(h)} = v_a] = \pi^{(h)}_a$), then
	$\pi^{(h)} \to  \pi^{(\infty)} = \left( \nicefrac 1n, \nicefrac 1n, \dotsc, \nicefrac 1n \right)^\top$,  as $h \to \infty$,
	and for all $h$,
	\begin{equation}\label{eq:prob-vector-convergence}
	\norm[1]{ \pi^{(h)} -  \pi^{(\infty)}} \le \sqrt{n} |\lambda_1|^h,
	\end{equation}
	where
	$| \lambda_1 | = \frac{1-p}{\sqrt{1+p^2 - 2p\cos \frac{2\pi}{n}}}$.
\end{lemma}

\begin{remark} \label{rem:lambda1}
For convenience, we also define the value $\lambda_1 \eqdef 0$ for \ssrandring\ (and any $0 \le p < 1$). With this notation, \cref{eq:prob-vector-convergence} holds in both cases.
\end{remark}

\begin{remark}
	For any probability vector $\pi$, it holds that
		$\norm[1]{\pi - \pi^{(\infty)}} \le \sqrt n$,
	and, thus, \cref{lem:markov-mixing} technically holds also for $h=0$.
\end{remark}

\begin{lemma}
  \label{lem:moments-of-gaussian-norm}
  Let %
  $N \sim \mathcal N(0, \sigma^2 I_d)$. Then,
    $\EE{\norm[2]{N}} 
    < \sigma \sqrt{d}$ and  %
   $\EE{\norm[2]{N}^2} 
	= d \sigma^2$.
\end{lemma}

\begin{lemma}[\hspace{-0.01cm}{\cite[Lem.~2]{Gubin1967}}]\label{lem:projections-norm}
  If the domain $\mathcal W \subset \Reals^d$ is convex and closed, then for any $\vect x, \vect y \in \Reals^d$, we have
    $\norm[2]{\vect x - \vect y} \ge \norm[2]{\Pi_\mathcal{W}(\vect x) - \Pi_\mathcal{W}(\vect y)}$.
\end{lemma}

\begin{lemma}\label{lem:norm-of-sum}
  For any $x,y \in \Reals^d$,
    $\norm[2]{x \pm y}^2 = \norm[2]{x}^2 + \norm[2]{y}^2 \pm 2 x^\top y$.
\end{lemma}

\subsection{Main Part of the Proof of \cref{thm:convergence}}
\label{sec:AppA_main_part}
We first consider the case of $H \geq 1$. 
For convenience, define 
\begin{align*}
  g_{h} & \triangleq \nabla f(\tau_{h-1}; \set D), \\
  \hat g_{h} & \triangleq \nabla f_{V^{(h)}} (\tau_{h-1}; \set D_{V^{(h)}}) + N_{h}
\end{align*} %
as a  shorthand notation for $h \in [H]$. With this notation, the token is updated as 
	$\tau_{h} \leftarrow \Pi_{\set W} \left(\tau_{h-1} - \eta_{h} \hat g_{h} \right)$.

If $V^{(h)}$ is  uniformly distributed over $\set V$, we have that $\EE{\hat g_{h}} = g_{h}$ for any fixed $\tau_{h-1}$, and in both schemes,
\begin{IEEEeqnarray*}{rCl}
  \EE{\norm[2]{\hat g_{h}}^2} &\stackrel{(a)}{=} &\EE{\norm[2]{\nabla f_{V^{(h)}} (\tau_{h-1}; \set D_{V^{(h)}})}^2} + \EE{\norm[2]{N_{h}}^2}
  \nonumber\\
  &&\,+\> 2 \EE{ N_{h}^\top \nabla f_{V^{(h)}} (\tau_{h-1}; \set D_{V^{(h)}})}
  \\
  & \stackrel{(b)}{=} &\EE{\norm[2]{\nabla f_{V^{(h)}} (\tau_{h-1}; \set D_{V^{(h)}})}^2} + \EE{\norm[2]{N_{h}}^2}
  \\
  & \stackrel{(c)}{\le} &k^2 + d \sigma^2,
\end{IEEEeqnarray*}
where $(a)$ is from \cref{lem:norm-of-sum}, $(b)$ is because $N_{h}$ is independent of other RVs and has zero mean, and $(c)$ follows from the $k$-Lipschitz property of $f$ and \cref{lem:moments-of-gaussian-norm}.

Now, we prove the main statement of \cref{thm:convergence}. In the proof, if it is not mentioned explicitly, the norm of a vector is the $\ell_2$-norm. Also, we assume the same dataset $\mathcal D$ everywhere and thus omit it for brevity.

Assume $H \ge 1$ is fixed (i.e., we condition on it). For any $\tau \in \set W$, by \cref{lem:projections-norm},
\begin{IEEEeqnarray*}{rCl}
  \IEEEeqnarraymulticol{3}{l}{%
    \EE{\norm{\Pi_{\set W}(\tau_{h-1} - \eta_{h} \hat g_{h}) - \Pi_{\set W}(\tau)}^2}}\nonumber\\*\hspace*{4.5cm}%
  & \le &\EE{\norm{\tau_{h-1} - \eta_{h} \hat g_{h} - \tau}^2}.\IEEEeqnarraynumspace
\end{IEEEeqnarray*}
Thus, %
\begin{IEEEeqnarray*}{rCl}
  \IEEEeqnarraymulticol{3}{l}{%
    \EE{\norm{\tau_{h} - \tau}^2}}\nonumber\\*%
  & = &\EE{\norm{\Pi_{\set W}(\tau_{h-1} - \eta_{h} \hat g_{h}) - \Pi_{\set W}(\tau)}^2}
  \\
  & \le &\EE{\norm{(\tau_{h-1} - \tau) - \eta_{h} \hat g_{h}}^2}
  \\
  & = &\EE{\norm{\tau_{h-1} - \tau}^2} + \eta_{h}^2 \EE{\norm{\hat g_{h}}^2} - 2 \eta_{h} \EE{(\tau_{h-1} - \tau)^\top \hat g_{h}}
  \\
  & \le &\EE{\norm{\tau_{h-1} - \tau}^2} + \eta_{h}^2 (k^2 + d \sigma^2)\! -\! 2 \eta_{h} \EE{(\tau_{h-1} - \tau)^\top \hat g_{h}}\\
  & \le &\EE{\norm{\tau_{h-1} - \tau}^2} - 2 \eta_{h} \EE{(\tau_{h-1} - \tau)^\top g_{h}}
  \nonumber\\
  && +\> \eta_{h}^2 (k^2 + d \sigma^2) + 2 \eta_{h} d_{\set W} k \sqrt{n} |\lambda_1|^{h},\IEEEeqnarraynumspace
\end{IEEEeqnarray*}
where the term $d_{\set W} k \sqrt{n} |\lambda_1|^{h}$ appears because of the difference between the distributions of $\hat g_h$ and $g_h$ (cf. \cref{lem:markov-mixing} and \cref{rem:lambda1}). Then,

\begin{IEEEeqnarray*}{rCl}
  \EE{(\tau_{h-1} - \tau)^\top g_{h}}& \le &\frac{\EE{\norm{\tau_{h-1} - \tau}^2}}{2 \eta_{h}} - \frac{\EE{\norm{\tau_{h} - \tau}^2}}{2 \eta_{h}} \nonumber\\
  && +\> \frac{\eta_{h} (k^2 + d \sigma^2)}{2} + d_{\set W} k \sqrt{n} |\lambda_1|^{h}.
\end{IEEEeqnarray*}

Let $j$ be an arbitrary element in $[H-1]$. Then, summing up and re-arranging, we get
\begin{IEEEeqnarray*}{rCl}
  \IEEEeqnarraymulticol{3}{l}{%
    \sum_{h = H - j}^{H} \EE{(\tau_{h-1} - \tau)^\top g_h}}\nonumber\\*\quad%
  & \le &\frac{\EE{\norm{\tau_{H-j-1} - \tau}^2}}{2 \eta_{H-j}}
  \nonumber\\
  && +\>\sum_{h=H-j}^{H-1} \frac{\EE{\norm{\tau_{h} - \tau}^2}}2 \left( \frac{1}{\eta_{h+1}} - \frac{1}{\eta_{h}} \right)  
  \nonumber\\
  && +\>\frac{k^2+d\sigma^2}{2} \sum_{h = H - j}^{H} \eta_{h} + d_{\set W} k \sqrt{n} \sum_{h = H - j}^{H} |\lambda_1|^{h}.
\end{IEEEeqnarray*}
Since $\tau_h, \tau \in \set W$, we have that $\norm{\tau_h - \tau}^2 \le d_{\set W}^2$. We also substitute $\eta_{h}$ with $\nicefrac{\zeta}{\sqrt{h}}$, which gives
\begin{IEEEeqnarray*}{rCl}
  \IEEEeqnarraymulticol{3}{l}{%
    \sum_{h = H - j}^{H} \EE{(\tau_{h-1} - \tau)^\top g_h}}\nonumber\\*\quad%
  & \le &\frac{\EE{\norm{\tau_{H-j-1} - \tau}^2} \sqrt{H-j}}{2 \zeta}
  + \frac{d_{\set W}^2}{2 \zeta} \left( \sqrt{H} - \sqrt{H-j} \right)
  \nonumber\\
  && +\> \frac{k^2+d\sigma^2}{2} \sum_{h = H - j}^{H} \frac{\zeta}{\sqrt{h}}+d_{\set W} k \sqrt{n} \sum_{h = H - j}^{H} |\lambda_1|^{h}.\IEEEeqnarraynumspace
\end{IEEEeqnarray*}
Here, we can upper bound the sum of inverse square roots 
as
\begin{IEEEeqnarray*}{rCl}
  \sum_{h=H-j}^{H} \frac{\zeta}{\sqrt{h}}& \le &\int_{H-j-1}^{H} \frac{\zeta}{\sqrt{h}} \mathrm d h
  =2 \zeta \left( \sqrt{H} - \sqrt{H-j-1} \right).  
\end{IEEEeqnarray*}%
Next, by convexity of $f$, we can lower bound $(\tau_{h-1} - \tau)^\top g_h$ by $f(\tau_{h-1}) - f(\tau)$. Hence, 
\begin{IEEEeqnarray*}{rCl}
  \IEEEeqnarraymulticol{3}{l}{%
    \sum_{h = H - j}^{H} \EE{f(\tau_{h-1}) - f(\tau)}}\nonumber\\*\quad%
  & \le &\sum_{h = H - j}^{H} \EE{(\tau_{h-1} - \tau)^\top g_h}
  \\
  & \le &\frac{\EE{\norm{\tau_{H-j-1} - \tau}^2} \sqrt{H-j}}{2 \zeta} + d_{\set W} k \sqrt{n} \sum_{h = H - j}^{H} |\lambda_1|^{h}
  \nonumber\\
  && +\>\frac{d_{\set W}^2}{2 \zeta} \left( \sqrt{H} - \sqrt{H-j} \right)  
  \nonumber\\
  && +\>\zeta (k^2+d\sigma^2) \left( \sqrt{H} - \sqrt{H-j-1} \right)
  \\
  & < &\frac{\EE{\norm{\tau_{H-j-1} - \tau}^2} \sqrt{H-j}}{2 \zeta} + d_{\set W} k \sqrt{n} \sum_{h = H - j}^{H} |\lambda_1|^{h} \nonumber\\
  && +\> \left( \frac{d_{\set W}^2}{2 \zeta} + \zeta (k^2+d\sigma^2) \right) \left( \sqrt{H} - \sqrt{H-j-1} \right)
  \\
  & = &\frac{\EE{\norm{\tau_{H-j-1} - \tau}^2} \sqrt{H-j}}{2 \zeta} + d_{\set W} k \sqrt{n} \sum_{h = H - j}^{H} |\lambda_1|^{h} \nonumber\\
  && +\> \left( \frac{d_{\set W}^2}{2 \zeta} + \zeta (k^2+d\sigma^2) \right) \frac{j+1}{\sqrt{H} + \sqrt{H-j-1}} \\	 
  & < &\frac{\EE{\norm{\tau_{H-j-1} - \tau}^2} \sqrt{H-j}}{2 \zeta} + d_{\set W} k \sqrt{n} \sum_{h = H - j}^{H} |\lambda_1|^{h} \\
  \nonumber\\
  && +\> \left( \frac{d_{\set W}^2}{2 \zeta} + \zeta (k^2+d\sigma^2) \right) \frac{j+1}{\sqrt{H}}.\IEEEyesnumber\IEEEeqnarraynumspace\label{eq:conv-proof-eq1}
\end{IEEEeqnarray*}
By setting $\tau = \tau_{H-j-1}$ in \eqref{eq:conv-proof-eq1}, we get
\begin{IEEEeqnarray*}{rCl}
  \IEEEeqnarraymulticol{3}{l}{%
    \sum_{h = H - j}^{H}\EE{f(\tau_{h-1}) - f(\tau_{H-j-1})}
  }\nonumber\\*\quad%
  & \le &\left( \frac{d_{\set W}^2}{2\zeta} + \zeta(k^2 + d\sigma^2) \right) \frac{j+1}{\sqrt{H}} 
  + d_{\set W} k \sqrt{n} \sum_{h = H - j}^{H} |\lambda_1|^{h}.
\end{IEEEeqnarray*}

Next, as a shorthand, let $S_j$ denote the average of the following $j+1$ iterates:
  $S_j = \frac{1}{j+1} \sum_{h = H - j}^{H} f(\tau_{h-1})$.
Then,
\begin{IEEEeqnarray*}{rCl}
  \IEEEeqnarraymulticol{3}{l}{%
    (j+1) \EE{S_j} - (j+1) \EE{f(\tau_{H-j-1})}
  }\nonumber\\*\quad%
  & = &\sum_{h = H - j}^{H} \EE{f(\tau_{h-1}) - f(\tau_{H-j-1})}
  \\
  & \le &\left( \frac{d_{\set W}^2}{2\zeta} + \zeta(k^2 + d\sigma^2) \right) \frac{j+1}{\sqrt{H}} 
  + d_{\set W} k \sqrt{n} \sum_{h = H - j}^{H} |\lambda_1|^{h}.\IEEEeqnarraynumspace
\end{IEEEeqnarray*}
Hence,
\begin{IEEEeqnarray*}{rCl}
  -\EE{f(\tau_{H-j-1})}& \le &- \EE{S_j}+ \left( \frac{d_{\set W}^2}{2\zeta} + \zeta(k^2 + d\sigma^2) \right) \frac{1}{\sqrt{H}}
  \nonumber\\
  && +\> \frac{d_{\set W} k \sqrt{n}}{j+1} \sum_{h = H - j}^{H} |\lambda_1|^{h} .
\end{IEEEeqnarray*}
Using this, we have
\begin{IEEEeqnarray*}{rCl}
  \EE{S_{j-1}}& = &\frac{(j+1)\EE{S_j} - \EE{f(\tau_{H-j-1})}}{j}
  \\
  & \le &
  \EE{S_j}+ \left( \frac{d_{\set W}^2}{2\zeta} + \zeta(k^2 + d\sigma^2) \right) \frac{1}{j\sqrt{H}} 
  \nonumber\\
  && +\>\frac{d_{\set W} k \sqrt{n}}{j(j+1)} \sum_{h = H - j}^{H} |\lambda_1|^{h}.\IEEEeqnarraynumspace
\end{IEEEeqnarray*}
In the following, to simplify notation, define
\begin{IEEEeqnarray*}{c}
  a_j \triangleq \left( \frac{d_{\set W}^2}{2\zeta} + \zeta(k^2 + d\sigma^2) \right) \frac{1}{j\sqrt{H}} 
  + \frac{d_{\set W} k \sqrt{n}}{j(j+1)} \sum_{h = H - j}^{H} |\lambda_1|^{h}\IEEEeqnarraynumspace
\end{IEEEeqnarray*}
as a shorthand. 
Then,
\begin{IEEEeqnarray*}{rCl}
  \EE{f(\tau_{H-1})}& = &\EE{S_0} \le \EE{S_1} + a_1 \le \EE{S_2} + a_1 + a_2
  \\
  & \le &\cdots \le \EE{S_{H-1}} + \sum_{j=1}^{H-1} a_j.
\end{IEEEeqnarray*}
Next, we bound a part of the sum on the right hand side as
\begin{multline*}
	\sum_{j=1}^{H-1} \left( \frac{d_{\set W}^2}{2\zeta} + \zeta(k^2 + d\sigma^2) \right) \frac{1}{j\sqrt{H}} \\
	\le \sum_{j=1}^{H-1} \left( \frac{d_{\set W}^2}{\zeta} + \zeta(k^2 + d\sigma^2) \right) \frac{1}{j\sqrt{H}} \\
	\le \frac{d_{\set W}^2 + \zeta^2(k^2 + d \sigma^2)}{\zeta \sqrt{H}} (1 + \log H) 
\end{multline*}
and obtain
\begin{IEEEeqnarray*}{rCl}
  \EE{f(\tau_{H-1})}& \le &\EE{S_{H-1}}
  + \frac{d_{\set W}^2 + \zeta^2(k^2 + d \sigma^2)}{\zeta \sqrt{H}} (1 + \log H)
  \nonumber\\
  && +\> \sum_{j=1}^{H-1} \frac{d_{\set W} k \sqrt{n}}{j(j+1)} \sum_{h = H - j}^{H} |\lambda_1|^{h}.\IEEEyesnumber\label{eq:something-1}\IEEEeqnarraynumspace
\end{IEEEeqnarray*}

Now, recall \eqref{eq:conv-proof-eq1}. Set there $j = H-1$ (i.e., $H-j=1$), $\tau = \tau^*$, and bound all norms by $d_{\set W}^2$, which results in
\begin{IEEEeqnarray*}{rCl}
  \IEEEeqnarraymulticol{3}{l}{%
    \sum_{h = 1}^{H} \EE{f(\tau_{h-1}) - f(\tau^*)}
  }\nonumber\\*%
  & \le &\frac{d_{\set W}^2}{2 \zeta} + d_{\set W} k \sqrt{n} \sum_{h = 1}^{H} |\lambda_1|^h 
  + \left( \frac{d_{\set W}^2}{2 \zeta} + \zeta (k^2+d\sigma^2) \right) \sqrt H
  \\
  & \le &\left( \frac{d_{\set W}^2}{\zeta} + \zeta (k^2+d\sigma^2) \right) \sqrt H +  d_{\set W} k \sqrt{n} \sum_{h = 1}^{H} |\lambda_1|^h.\IEEEeqnarraynumspace
\end{IEEEeqnarray*}
Therefore,
\begin{IEEEeqnarray*}{rCl}
  \IEEEeqnarraymulticol{3}{l}{%
    \EE{S_{H-1}} - f(\tau^*) %
   = \EE{\frac 1H \sum_{h = 1}^{H} \left( f(\tau_{h-1}) - f(\tau^*) \right)}}
  \\
  & \le &\frac{d_{\set W}^2 + \zeta^2(k^2 + d \sigma^2)}{\zeta \sqrt H} +  \frac{d_{\set W} k \sqrt{n}}{H} \sum_{h = 1}^{H} |\lambda_1|^h.\IEEEyesnumber\IEEEeqnarraynumspace\label{eq:something-2}
\end{IEEEeqnarray*}
Finally, by combining \eqref{eq:something-1} and \eqref{eq:something-2}, we obtain
\begin{IEEEeqnarray*}{rCl}
  \IEEEeqnarraymulticol{3}{l}{%
    \EE{f(\tau_{H-1}) - f(\tau^*)}}\nonumber\\*[1mm]\,%
  & \le &\frac{(d_{\set W}^2 + \zeta^2(k^2 + d \sigma^2))(2 + \log H)}{\zeta \sqrt H}
  \nonumber\\
  && +\> d_{\set W} k \sqrt n \Biggl( \frac{1}{H} \sum_{h=1}^{H} |\lambda_1|^h + \sum_{j=1}^{H-1} \frac{1}{j(j+1)} \sum_{h = H - j}^{H} |\lambda_1|^h \Biggr).\IEEEeqnarraynumspace
\end{IEEEeqnarray*}
Then,
\begin{IEEEeqnarray*}{rCl}
	\IEEEeqnarraymulticol{3}{l}{%
		\EE{f(\tau_{H}) - f(\tau^*)}}\nonumber\\*[1mm]\,%
	& \le &\frac{(d_{\set W}^2 + \zeta^2(k^2 + d \sigma^2))(2 + \log (H+1))}{\zeta \sqrt{H+1}}
	\nonumber\\
	&& \hspace{-0.6cm}+\> d_{\set W} k \sqrt n \Biggl( \frac{1}{H+1} \sum_{h=1}^{H+1} |\lambda_1|^h + \sum_{j=1}^{H} \frac{1}{j(j+1)} \sum_{h = H - j + 1}^{H+1} |\lambda_1|^h \Biggr).
	\IEEEeqnarraynumspace
\end{IEEEeqnarray*}

The corner case of $H=0$ (and thus, $\tau_{h_{\max}} = \tau_0 = 0$) can be bounded as %
  $|f(0) - f(\tau^*)| \le k \norm{0 - \tau^*} \le k d_{\set W}$.

As a final step, we need to take expectation conditioned on the distribution of $H$, which is binomial with $h_{\max}$ independent trials and success probability $1-p$, i.e.,
\[
  \Pr[H=h] = \binom{h_{\max}}{h}(1-p)^h p^{h_{\max} - h},
\]
which concludes the proof.

\appendices
\setcounter{section}{1}  
\section{Proof of \cref{thm:ss_ring_all_noise,thm:ss_rand_ring_all_noise}}
\label{app:AppB}

The main tool of the proofs is the concept of privacy amplification by iteration \cite{FeldmanMironovTalwarThakurta18_1}, and  Theorem~22 therein. %
The setting in \cite{FeldmanMironovTalwarThakurta18_1}  is \emph{projected noisy stochastic gradient decent}, in which noise is added for every gradient update step. %
The main technical tool is R{\'e}nyi divergence and the proof evolves around upper bounding it for a single view of a node. In particular, based on \cref{lem:recursion}, for any distinct pair of users $u,v$, we can derive an upper bound on the R{\'e}nyi divergence between the views of user $v$ when the token visits for the $(r+1)$-th time, excluding received and sent messages observed up to and including the $r$-th visit, for two neighboring datasets of user $u$ (\cref{lem:gen_theorem23}). By maximizing this upper bound over all pairs of distinct users $u,v$ and 
by using a composition theorem for RDP \cite[Prop.~1]{Mironov17_1} (\cref{lem:summation-bounds_RDiv}), %
we can derive an upper bound on the RDP level of \cref{alg:ss}, which can be transformed  into an upper bound on the DP level using \cite[Prop.~3]{Mironov17_1} (\cref{lem:RDPtoDP}). %
In order to get the best (lowest) upper bound, the R{\'e}nyi divergence parameter $\alpha$ can be optimized. 
Finally, since the number of visits to a node is not a constant, but instead follows a binomial distribution, a standard Chernoff bound in combination with \cref{lem:RandomComp} can be used to derive the final result. }    %

We start by defining R{\'e}nyi divergence and RDP %
and then state some important results from the privacy amplification by iteration literature.   In particular, definitions and results from \cite{Mironov17_1,FeldmanMironovTalwarThakurta18_1}.

\subsection{Important Results From \cite{Mironov17_1,FeldmanMironovTalwarThakurta18_1}}
\label{sec:results-from-PA_iteration}
We start by stating and adapting some important definitions and results from \cite{Mironov17_1,FeldmanMironovTalwarThakurta18_1}. Central to the arguments in \cite{FeldmanMironovTalwarThakurta18_1} is the concept of R{\'e}nyi divergence and shifted R{\'e}nyi divergence.

\begin{definition}[R{\'e}nyi divergence]
  \label{def:Renyi-divergence}
  For two probability distributions $\mu$ and $\nu$ defined over the same  set $\set{Z}$, the R{\'e}nyi divergence of  positive order $\alpha\neq 1$ between $\mu$ and $\nu$  is
  \begin{IEEEeqnarray*}{c}    
    \RDiv[\alpha]{\mu}{\nu}\eqdef\frac{1}{\alpha-1}\log{\int_{z\in\set{Z}}\left(\frac{\mu(z)}{\nu(z)}\right)^{\alpha}\nu(z)\dd z}.
  \end{IEEEeqnarray*}
\end{definition}

\begin{definition}[{Shifted R{\'e}nyi divergence~\cite[Def.~8]{FeldmanMironovTalwarThakurta18_1}}]
  \label{def:shifted-Renyi-divergence}
  For two probability distributions $\mu$ and $\nu$ defined over the same  complete normed vector space $(\set{Z},\enorm{}{\cdot})$, the $u$-shifted R{\'e}nyi divergence, for $u \geq 0$, of order $\alpha> 1$ between $\mu$ and $\nu$ is
  \begin{IEEEeqnarray*}{c}
    \sRDiv[\alpha]{u}{\mu}{\nu}\eqdef\inf_{\mu'\colon \dW(\mu,\mu')\leq u}\RDiv[\alpha]{\mu'}{\nu},
  \end{IEEEeqnarray*}
  where $\dW(\cdot,\cdot)$ denotes the $\infty$-Wasserstein distance~\cite[Def.~6]{FeldmanMironovTalwarThakurta18_1} between two distributions on $(\set{Z},\enorm{}{\cdot})$.
\end{definition}

\begin{lemma}[{Weak convexity R{\'e}nyi divergence~\cite[Lem.~25]{FeldmanMironovTalwarThakurta18_1}}]
\label{lem:weak-convexity-Renyi-divergence}
Let $\mu_1,\ldots,\mu_n$ and $\nu_1,\ldots,\nu_n$ be probability distributions defined on a complete normed vector space $(\set{Z},\enorm{}{\cdot})$ such that $\forall\, i \in [n]$, $\RDiv[\alpha]{\mu_i}{\nu_i} \leq \nicefrac{b}{(\alpha-1)}$ for some $b \in (0,1]$ where $\alpha>1$. Let $\rho$ be a  probability distribution over $[n]$ and denote by $\mu_\rho$ the probability distribution over $\set Z$ obtained by sampling $i$ from $\rho$ and then outputting a random sample from $\mu_i$ (respectively, $\nu_i$). Then
 \begin{IEEEeqnarray*}{c}
 \RDiv[\alpha]{\mu_\rho}{\nu_\rho} \leq (1+b) \cdot \mathbb{E}_{i \sim \rho} \RDiv[\alpha]{\mu_i}{\nu_i}.
  \end{IEEEeqnarray*}
\end{lemma}

\begin{definition}[\hspace{-0.01cm}{\cite[Def.~10]{FeldmanMironovTalwarThakurta18_1}}]
  \label{def:supRDiv_shifted-distribuion}
  For a distribution $\zeta$ over $(\set{Z},\enorm{}{\cdot})$ and any $a\geq 0$, the \emph{magnitude of noise} is the largest R{\'e}nyi divergence of positive order $\alpha \neq 1$ between $\zeta$ and the same distribution $\zeta$ shifted by a vector of length at most $a$, i.e.,
  \begin{IEEEeqnarray*}{c}
    \supRDiv_\alpha(\zeta,a)\eqdef\sup_{{z}\colon\enorm{}{{z}}\leq a}\RDiv[\alpha]{\conv{\zeta}{{z}}}{\zeta}.
  \end{IEEEeqnarray*}
\end{definition}
\begin{remark}
  \label{rem:remark_Gaussian-distribution}
  Consider the standard Gaussian distribution over $\Reals^d$ with variance $\sigma^2$, denoted by $\eNormal{0}{\sigma^2 I_d}$. Then, it is known that $\forall\,z\in\Reals^d,\sigma>0$ (see, e.g.,~\cite[Ex.~3]{vanErvenHarremoes14_1}), we have
  \begin{equation*}
    \begin{IEEEeqnarraybox}[
      \IEEEeqnarraystrutmode
      \IEEEeqnarraystrutsizeadd{2pt}{2pt}][c]{rCl}
      \RDiv[\alpha]{\Normal{x}{\sigma^2 I_d}}{\Normal{0}{\sigma^2 I_d}}& = &\alpha\frac{\enorm{}{x}^2}{2\sigma^2},
      \\[1mm]
      \supRDiv_\alpha\bigl(\Normal{0}{\sigma^2 I_d},a\bigr)& = &\alpha\frac{a^2}{2\sigma^2}.
    \end{IEEEeqnarraybox}
  \end{equation*}
\end{remark}

\begin{definition}[{Contractive noisy iteration (CNI)~\cite[Def.~19]{FeldmanMironovTalwarThakurta18_1}}]
  \label{def:contractive-noisy-iteration}
  Given an initial random state $Z_0\in\set{Z}$, a sequence of contractive maps $\{\psi_h\}_{h=1}^m$, and a sequence of noise distributions $\{\zeta_h\}_{h=1}^m$, the \emph{contractive noisy iteration} after $m$ steps, denoted by $\mathrm{CNI}_m\bigl(Z_0,\{\psi_h\},\{\zeta_h\}\bigr)$, is defined by the following update process:
    $Z_{h}\eqdef\psi_{h}(Z_{h-1})+N_{h}$,
  where $N_{h}\sim\zeta_{h}$, $h\in[m]$.
\end{definition}

The following lemma is taken from~\cite[Thm.~22]{FeldmanMironovTalwarThakurta18_1}.
\begin{lemma}[{\hspace{-0.01cm}\cite[Thm.~22]{FeldmanMironovTalwarThakurta18_1}}]
  \label{lem:recursion}
  Let $Z_m$ and $Z'_m$ represent the outputs of $\mathrm{CNI}_m\bigl(Z_0,\{\psi_h\},\{\zeta_h\}\bigr)$ and $\mathrm{CNI}_m\bigl(Z_0,\{\psi'_h\},\{\zeta_h\}\bigr)$, respectively. Define $s_h\eqdef\sup_{z}\bignorm{\psi_h(z)-\psi'_h(z)}$, $\{a_h\}_{h=1}^m$ a sequence of nonnegative reals, and $u_h\eqdef\sum_{i=1}^h (s_i-a_i)$. If $u_h\geq 0$, $\forall\,h\in[m]$, %
  then
    $\sRDiv[\alpha]{u_m}{Z_m}{Z'_m}\leq\sum_{h\in[m]}\supRDiv_{\alpha}(\zeta_h,a_h)$.
\end{lemma}

Now, we review some results from %
RDP~\cite{Mironov17_1}.
\begin{definition}[$(\alpha,\veps)$-RDP]
  \label{def:def_RDP}
  For any positive $\alpha \neq 1$ and $\varepsilon\geq 0$, a (randomized) protocol $\set{A}$ is said to satisfy  $(\alpha,\veps)$-RDP, if for all neighboring datasets $\set{D},\set{D}'$ and for all $\set{S}$ in the output space $\Omega$, we have
    $\RDiv[\alpha]{\set{A}(\set{D})\in\set{S}}{\set{A}(\set{D}')\in\set{S}}\leq \veps$.
\end{definition}

Next, we state the composition theorem for RDP.
\begin{lemma}[\hspace{-0.01cm}{\cite[Prop.~1]{Mironov17_1}}]
  \label{lem:summation-bounds_RDiv}
  Let $r\in\Naturals$. If $\{\set{A}_l\}_{l=1}^{r}$ are protocols satisfying, respectively, $(\alpha,\veps_1)$-RDP, $\ldots$, $(\alpha,\veps_r)$-RDP, then their composition defined as $(\set{A}_1,\ldots,\set{A}_r)$ satisfies $(\alpha,\sum_{i=1}^r \veps_i)$-RDP.
\end{lemma}

The DP (RDP) level with a random number of entries in the composition can be bounded as follows.
\begin{lemma}
  \label{lem:RandomComp}
  Let $R$ denote a RV with range $\{1,2,\ldots\}$ that satisfies $\Prs{R>r} \leq \delta'$. If $\{\set{A}_l\}_{l=1}^{R}$ are protocols satisfying, respectively, $(\veps_1,\delta_1)$-DP, $\ldots$, $(\veps_R,\delta_R)$-DP, then their composition defined as $(\set{A}_1,\ldots,\set{A}_R)$ satisfies $(\veps_\mathrm{c}, \delta_\mathrm{c} + \delta')$-DP, where $(\veps_c,\delta_c)$ is the  DP guarantee under $r$-fold composition for DP.
\end{lemma}

In particular, if $R$ is a binomial RV (i.e., a sum of independent Bernoulli RVs), we can use the standard Chernoff bound to upper bound $\Prs{R>r}$.

A relation between $(\alpha,\veps)$-RDP and $(\veps,\delta)$-DP can be stated as follows.
\begin{lemma}[\hspace{-0.01cm}{\cite[Prop.~3]{Mironov17_1}}] \label{lem:RDPtoDP}
  If $\set{A}$ satisfies $(\alpha,\veps)$-RDP for $\alpha >1$, then for all $\delta\in(0,1)$, it also satisfies $\Bigl(\veps+\frac{\log{(\nicefrac{1}{\delta})}}{\alpha-1},\delta\Bigr)$-DP.
\end{lemma}

\subsection{Adapting to \cref{alg:ss}} %

For notational convenience, let  $\set O^{(r)}_v(\set A(\set D))$ be the view of user $v$ when the token visits for the $r$-th time, excluding sent/received messages observed up to and including the $(r-1)$-th visit.

The following lemma is analogous to~\cite[Thm.~23]{FeldmanMironovTalwarThakurta18_1}, but tailored to our setting with a decreasing learning rate.

\begin{lemma} \label{lem:gen_theorem23}
Let $\set W \subseteq \Reals^d$ be a convex set and let $f_v: \set W \times \set R^\kappa \rightarrow \Reals$, $v \in \set V$, be $k$-Lipschitz continuous and $\beta$-smooth convex functions in their first argument. Let $(v_1^{(r+1)},\ldots,v_{l^{(r+1)}}^{(r+1)})$ denote the sequence of nodes visited in between the $r$-th and $(r+1)$-th visit to node $v$ in \cref{alg:ss}. %
Then, for \cref{alg:ss} with learning rate parameter $0 < \zeta \leq \nicefrac{2}{\beta}$ and constant noise $\sigma_h = \sigma$, and any distinct pair of users $u,v \in \set V$, 
\begin{IEEEeqnarray*}{rCl} 
\IEEEeqnarraymulticol{3}{l}{\RDiv[\alpha]{\set O^{(r+1)}_v(\set A(\set D))}{\set O^{(r+1)}_v(\set A(\set D'))}}\nonumber\\
& & \begin{cases} 
 \leq \frac{2\alpha k^2}{\sigma^2} %
\qquad \quad  \text{if $\xi_{u,v}^{(r+1)}=1$}, &\\ 
\leq \frac{\alpha k^2\xi_{u,v}^{(r+1)}}{2 \left(1 + \sum_{i=1}^r \xi_{u,v}^{(i)}\right) \cdot \left(\sqrt{1 + \sum_{i=1}^r \xi_{u,v}^{(i)} + \xi_{u,v}^{(r+1)}} - \sqrt{1 + \sum_{i=1}^r \xi_{u,v}^{(i)}}\right)^2 \sigma^2} & \\
\qquad \qquad \qquad %
\text{if $1 < \xi_{u,v}^{(r+1)} < \infty$}, & \\ %
=0 %
\qquad \qquad\; \text{otherwise}, & \end{cases}
\end{IEEEeqnarray*}
for every $\alpha > 1$, where $\set D \sim_u \set D'$, $\xi_{u,v}^{(r+1)} \eqdef l^{(r+1)} -c^{(r+1)} +1$ %
and $c^{(r+1)} \in [l^{(r+1)}]$ is the index of $v_i^{(r+1)}=u$ for $u \in \{v_1^{(r+1)},\ldots,v_{l^{(r+1)}}^{(r+1)}\}$, i.e., %
$u = v_{c^{(r+1)}}^{(r+1)}$. Otherwise, if $u \not\in \{v_1^{(r+1)},\ldots,v_{l^{(r+1)}}^{(r+1)}\}$, then $\xi_{u,v}^{(r+1)} \eqdef \infty$. %
\end{lemma}

For simplicity of notation, we omit the superscript $(r+1)$ from $l$, $c$, and $v_1,\ldots,v_l$ in the following.

\begin{IEEEproof}
Consider the case when $u \in \{v_1,\ldots,v_l\}$. Otherwise, $\set O^{(r+1)}_v(\set A(\set D)) = \set O^{(r+1)}_v(\set A(\set D'))$, and it follows directly that $\RDiv[\alpha]{\set O^{(r+1)}_v(\set A(\set D))}{\set O^{(r+1)}_v(\set A(\set D'))}=0$.

By assumption, the learning rate  $\eta_{h_i}$ is upper-bounded by $\nicefrac{2}{\beta}$, and hence the update rule  $g^{(v)}_r(\tau; \mathtt{state}_v(h))$ in %
\eqref{eq:g_update} for \cref{alg:ss} constitutes a CNI (see \cite[Prop.~18]{FeldmanMironovTalwarThakurta18_1}).
Consider now the CNI from \cref{def:contractive-noisy-iteration} with $\psi_i(\tau) = \prod_{\set W}( \tau - \eta_{h_i} \nabla f_{v_i}(\tau,\set{D}_{v_i})) =  \prod_{\set W}( \tau) - \eta_{h_i} \nabla f_{v_i}(\prod_{\set W}(\tau),\set{D}_{v_i})$ and with $\psi'_h(\tau) = \prod_{\set W}( \tau - \eta_{h_i} \nabla f_{v_i}(\tau,\set{D}'_{v_i})) =  \prod_{\set W}( \tau) - \eta_{h_i} \nabla f_{v_i}(\prod_{\set W}(\tau),\set{D}'_{v_i})$, corresponding to $g^{(v)}_r(\tau; \mathtt{state}_v(h))$ in  \eqref{eq:g_update}. %
It follows that
\begin{IEEEeqnarray*}{rCl}
\IEEEeqnarraymulticol{3}{l}{\sup_\tau \bigl\lVert \psi_i(\tau) -\psi'_i(\tau) \bigr\rVert_2}\\
 & =  &\sup_\tau \left\lVert  \eta_{h_i} \nabla f_{v_i}\left(\prod_{\set W}(\tau),\set{D}_{v_i}\right) -  \eta_{h_i} \nabla f_{v_i}\left(\prod_{\set W}(\tau),\set{D}'_{v_i}\right) \right\rVert_2 \\
 & = & \begin{cases} 0 & \text{if $i \neq c$}, \\
\leq 2 \eta_{h_c} k & \text{otherwise}, \end{cases}
\end{IEEEeqnarray*}
since by assumption $f_{v_i}$ is $k$-Lipschitz continuous. 

Now apply \cref{lem:recursion} with $a_i = 0$, $\forall\, i \in [c-1]$, and $a_i = \nicefrac{2 \eta_{h_i} k}{\varrho_{u,v}^{(r+1)}}$, $\forall\, i \in [c:l]$, where %
\begin{IEEEeqnarray}{c}
\varrho_{u,v}^{(r+1)} \triangleq \frac{\sum_{i \in [c:l]} %
\eta_{h_i}}{\eta_{h_c}} = \frac{\sum_{i \in [c:l]} %
\frac{1}{\sqrt{h_i}}}{\frac{1}{\sqrt{h_c}}} \label{eq:varrho_uv}.
\end{IEEEeqnarray}
Clearly, $z_i = s_i-a_i \geq 0$, $\forall\, i \in [l]$, and $z_l = 0$. Hence, using \cref{rem:remark_Gaussian-distribution},
\begin{IEEEeqnarray}{rCl}
\IEEEeqnarraymulticol{3}{l}{\RDiv[\alpha]{\set O^{(r+1)}_v(\set A(\set D))}{\set O^{(r+1)}_v(\set A(\set D'))} } \nonumber \\
\qquad & \leq & \alpha \sum_{i \in [c:l]} %
\frac{4\eta_{h_i}^2 k^2}{2 \left(\varrho_{u,v}^{(r+1)}\right)^2 \eta_{h_i}^2 \sigma^2}
= \alpha \sum_{i \in [c:l]} %
\frac{2 k^2}{\left(\varrho_{u,v}^{(r+1)}\right)^2 \sigma^2} \nonumber \\
 \qquad & = &   \frac{2 \alpha |[c:l]| %
k^2}{\left(\varrho_{u,v}^{(r+1)}\right)^2 \sigma^2} 
= \frac{2 \alpha  \xi_{u,v}^{(r+1)}  k^2}{\left(\varrho_{u,v}^{(r+1)}\right)^2 \sigma^2}. \label{eq:div_alpha_expression}
\end{IEEEeqnarray}
Now, if $c=l$, i.e., $u=v_l$ and $\xi_{u,v}^{(r+1)}=1$, then from \eqref{eq:varrho_uv} it follows that $\varrho_{u,v}^{(r+1)}=1$ and therefore %
\[\RDiv[\alpha]{\set O^{(r+1)}_v(\set A(\set D))}{\set O^{(r+1)}_v(\set A(\set D'))} \leq \frac{2\alpha k^2}{\sigma^2}.\] Otherwise, i.e., when $l > c$ and $1 < \xi_{u,v}^{(r+1)} < \infty$,
\begin{IEEEeqnarray}{rCl}
\varrho_{u,v}^{(r+1)} & = &\frac{\sum_{i \in [c:l]} %
\frac{1}{\sqrt{h_i}}}{\frac{1}{\sqrt{h_c}}} \nonumber \\ %
& \overset{(a)}{>} & 2 \sqrt{h_c} \left(\sqrt{h_{c} + \xi_{u,v}^{(r+1)}-1 +1} -\sqrt{h_{c}}\right) %
 \label{eq:varrho_uv_third}\\
& \overset{(b)}{\geq} & 2\sqrt{1 + \sum_{i=1}^r \xi_{u,v}^{(i)}} \cdot \left(\sqrt{1 +  \sum_{i=1}^r \xi_{u,v}^{(i)} + \xi_{u,v}^{(r+1)}} \right.  \nonumber \\
& & \qquad \qquad  \qquad \qquad \quad - \left. \sqrt{1 +  \sum_{i=1}^r \xi_{u,v}^{(i)}}\right), \nonumber
\end{IEEEeqnarray}
where $(a)$ follows by taking the anti-derivative of $\nicefrac{1}{\sqrt{h_i}}$ and the fact that the learning rate is only updated when visiting a node, i.e., $h_{l} = h_{l-1}+1 = h_{l-2}+2=\cdots=h_{c}+l-c$, and $(b)$ follows by lower-bounding $h_c$ by $1 +  \sum_{i=1}^r \xi_{u,v}^{(i)}$ (the expression in \eqref{eq:varrho_uv_third} is strictly increasing in $h_c$ for $\xi_{u,v}^{(r+1)} > 0$). In particular, for $r=0$, $h_c \geq 1$, which is obviously true. For $r=1$ (the second visit), the token has at least made $\xi_{u,v}^{(1)}$ updates, etc., from which the lower bound on $h_c$ follows.
\end{IEEEproof}

\subsection{Proof of \cref{thm:ss_ring_all_noise}}

For the \ssring\ scheme, in every round $r$ (unless all nodes are skipped), there exists a pair of neighboring nodes $(\tilde{u}^{(r)},\tilde{v}^{(r)})$ for which the token travels directly from $\tilde{u}^{(r)}$ to $\tilde{v}^{(r)}$. Hence, $\xi_{{\tilde{u}^{(r)},\tilde{v}}^{(r)}}^{(r)} = 1$ for all $r$, and
it follows from \cref{lem:gen_theorem23} that
  \begin{IEEEeqnarray}{rCl} 
\hspace{-2ex}\max_{u,v \in \set V,\ u \neq v} \RDiv[\alpha]{\set O_v^{(r)}(\set A(\set D))}{\set O_v^{(r)}(\set A(\set D'))}  %
  & \leq & \frac{2\alpha k^2}{ \sigma^2}. %
 \label{eq:maxRenyiDivDetRing}
 \end{IEEEeqnarray}
The number of visits of the token to a node $v$ during the execution of the algorithm, denoted by $\Xi_v$, follows a binomial distribution with parameters $\nicefrac{h_{\max}}{n}$ (number of independent trials) and $1-p$ (success probability). Let $\tilde{h}$ be defined as in the formulation of the theorem. Then, it follows from a standard Chernoff bound that ${\mathrm{Pr}}(\Xi_v \geq \tilde{h}) \leq \delta'$, for some $\delta' \in (0,1)$. Now,
\begin{IEEEeqnarray}{rCl} 
 \IEEEeqnarraymulticol{3}{l}{\max_{u,v \in \set V,\ u \neq v} \RDiv[\alpha]{\set O_v(\set A(\set D))}{\set O_v(\set A(\set D'))}}  \nonumber \\ \qquad
  & \overset{(a)}{\leq} & \max_{u,v \in \set V,\ u \neq v} \sum_{r=1}^{\tilde{h}}  \RDiv[\alpha]{\set O^{(r)}_v(\set A(\set D))}{\set O^{(r)}_v(\set A(\set D'))}  \nonumber \\
 & \overset{(b)}{\leq} & \sum_{r=1}^{\tilde{h}} \max_{u,v \in \set V,\ u \neq v} \RDiv[\alpha]{\set O^{(r)}_v(\set A(\set D))}{\set O^{(r)}_v(\set A(\set D'))}  \nonumber \\
& \overset{(c)}{\leq} & %
\sum_{r=1}^{\tilde{h}} \frac{2\alpha k^2}{\sigma^2} %
 =  \frac{2 \alpha k^2}{\sigma^2} \cdot \tilde{h} \label{eq:composition}
\end{IEEEeqnarray}
for every $\alpha > 1$, where $\set D \sim_u \set D'$. $(a)$ follows from  the composition theorem for RDP (\cref{lem:summation-bounds_RDiv}) and \cref{lem:RandomComp}, $(b)$ from swapping the order of maximization and summation, and $(c)$ from \eqref{eq:maxRenyiDivDetRing}.

Then, converting from RDP to DP using \cref{lem:RDPtoDP} gives that \cref{alg:ss} satisfies
\begin{IEEEeqnarray}{c} \label{eq:ring_single_noise_ndp}
\left( \frac{2 \tilde{h} \alpha k^2}{\sigma^2} + \frac{\log(\nicefrac{1}{\delta})}{\alpha-1}, \delta+\delta' \right)-\text{NDP}.
\end{IEEEeqnarray}

Now, the R{\'e}nyi divergence parameter $\alpha$ can be optimized in order to minimize $\nicefrac{2 \tilde{h} \alpha k^2}{\sigma^2} + \nicefrac{\log(\nicefrac{1}{\delta})}{(\alpha-1)}$ by taking the derivative with respect to $\alpha$. Doing so, gives
$\alpha = 1 + \frac{\sigma \sqrt{\log(\nicefrac{1}{\delta})}}{k \sqrt{2 \tilde{h}}} > 1$
from which the result follows by substituting this value of $\alpha$ 
into \eqref{eq:ring_single_noise_ndp} and setting $\sigma= \frac{k \sqrt{8 \log (\nicefrac{1.25}{\delta})}}{\varepsilon}$, where $\varepsilon > 0$  and $0 < \delta < 1$.

\subsection{Proof of \cref{thm:ss_rand_ring_all_noise}}

In contrast to the proof of \cref{thm:ss_ring_all_noise}, the distance between any pair of two nodes $u,v$ is random over the rounds of the algorithm. Hence, we have to resort to a  weak form of convexity for R{\'e}nyi divergence as formulated in \cref{lem:weak-convexity-Renyi-divergence}. 
We start with a technical lemma.
\begin{lemma}  \label{lem:technical_xi}
The fraction  $\nicefrac{\xi_{u,v}^{(r+1)}}{\left(\varrho_{u,v}^{(r+1)}\right)^2}$ from %
\eqref{eq:div_alpha_expression} is upper-bounded by $1$.
\end{lemma}

Now, let $\Xi_{u,v}^{(r)}$ denote the actual number of noise terms added in between the $(r-1)$-th and $r$-th visit of the token at node $v$ after visiting node $u$.
$\Xi_{u,v}^{(r)}$ is a binomial RV with parameters $d^{(r)}(u,v)$ and $1-p$, where $d^{(r)}(u,v)$ is the distance between $u$ and $v$ along the direction of the token over the ring. %
From \cref{lem:weak-convexity-Renyi-divergence}, it follows that 
\begin{IEEEeqnarray*}{rCl} 
\IEEEeqnarraymulticol{3}{l}{\RDiv[\alpha]{\set O^{(r)}_v(\set A(\set D))}{\set O^{(r)}_v(\set A(\set D'))} \leq (1+b)} \\
 & \times  &\mathbb{E} \left[ \RDiv[\alpha]{\set O^{(r)}_{v}(\set A(\set D))}{\set O^{(r)}_{v}(\set A(\set D')) \Bigl\vert \Xi_{u,v}^{(i)}=\xi_{u,v}^{(i)}, i \in [r]} \right],
\end{IEEEeqnarray*}
where $\RDiv[\alpha]{\set O^{(r)}_{v}(\set A(\set D))}{\set O^{(r)}_{v}(\set A(\set D')) \Bigl\vert \Xi_{u,v}^{(i)}=\xi_{u,v}^{(i)}, i \in [r]}$ is the R{\'e}nyi divergence between the views $\set O^{(r)}_{v}(\set A(\set D))$ and $\set O^{(r)}_{v}(\set A(\set D'))$ given that in between the $(i-1)$-th and $i$-th visit of the token at node $v$, $\xi_{u,v}^{(i)} \in  [d^{(i)}(u,v)]$ nodes after node $u$ (including) have been visited,
and where $0 < b \leq 1$ is a constant such that 
\begin{IEEEeqnarray}{rCl}
  \IEEEeqnarraymulticol{3}{l}{%
    \RDiv[\alpha]{\set O^{(r)}_{v}(\set A(\set D))}{\set O^{(r)}_{v}(\set A(\set D')) \Bigl\vert \Xi_{u,v}^{(i)}=\xi_{u,v}^{(i)}, i \in [r]} }\nonumber\\*\quad%
  & \leq & \frac{b}{\alpha-1}
\label{eq:b_condition}
\end{IEEEeqnarray}
for all $\xi_{u,v}^{(i)} \in [d^{(i)}(u,v)]$. By picking $b=1$ and  applying \cref{lem:gen_theorem23}, %
 gives the expression in \eqref{eq:maxRenyiDivRndRingDecLearnRate} at the top of the %
 next page. 
\begin{figure*}[t]
		\normalsize
		\begin{IEEEeqnarray}{rCl} 
			\IEEEeqnarraymulticol{3}{l}{\max_{u,v \in \set V,\ u \neq v} \RDiv[\alpha]{\set O^{(r)}_v(\set A(\set D))}{\set O^{(r)}_v(\set A(\set D'))}}\nonumber\\
			\quad &  \leq \max_{u,v \in \set V,\ u \neq v} (1+1) &\cdot \frac{2\alpha k^2}{\sigma^2} \mathbb{E}_{\Xi_{u,v}^{(1)},\ldots,\Xi_{u,v}^{(r)}}\left[\frac{\Xi_{u,v}^{(r)}}{4 (1 +\sum_{i=1}^{r-1} \Xi_{u,v}^{(i)}) \cdot \left(\sqrt{1 + \sum_{i=1}^{r-1} \Xi_{u,v}^{(i)} + \Xi_{u,v}^{(r)}} - \sqrt{1 + \sum_{i=1}^{r-1} \Xi_{u,v}^{(i)}}\right)^2  } \right] \label{eq:maxRenyiDivRndRingDecLearnRate}
		\end{IEEEeqnarray}
		\hrulefill
		\vspace*{-2mm}
\end{figure*} %
 As $\nicefrac{\xi_{u,v}^{(r+1)}}{\left(\varrho_{u,v}^{(r+1)}\right)^2} \leq 1$ (see \cref{lem:technical_xi}), in order to satisfy \eqref{eq:b_condition} (with $b=1$), we require that 
$2\alpha (\alpha-1)k^2 \leq \sigma^2$ (see \eqref{eq:div_alpha_expression}), which is equivalent to
$\frac{1-\sqrt{2\frac{\sigma^2}{k^2}+1}}{2} \leq \alpha \leq \frac{1+\sqrt{2\frac{\sigma^2}{k^2}+1}}{2}$.
Since the lower bound on $\alpha$ above is less than one, 
\begin{IEEEeqnarray}{c}
1 < \alpha \leq \frac{1+\sqrt{2\frac{\sigma^2}{k^2}+1}}{2} = \frac{1+\sqrt{\frac{16  \log (\nicefrac{1.25}{\delta})}{\varepsilon^2} +1 }}{2},
 \label{eq:condition_alpha}
 \end{IEEEeqnarray}
where we have used that $\sigma= \frac{k \sqrt{8 \log (\nicefrac{1.25}{\delta})}}{\varepsilon}$.

In the following, to simplify notation, let $g\bigl(\Xi_{u,v}^{(1)}\ldots,\Xi_{u,v}^{(r)} \bigr)$ denote the expression inside the expectation operator of \eqref{eq:maxRenyiDivRndRingDecLearnRate}. 
It follows that
\begin{IEEEeqnarray}{rCl}
\IEEEeqnarraymulticol{3}{l}{\mathbb{E}_{\Xi_{u,v}^{(1)},\ldots,\Xi_{u,v}^{(r)}}\left[g\bigl(\Xi_{u,v}^{(1)}\ldots,\Xi_{u,v}^{(r)} \bigr) \right]}\nonumber \\
  & = & \sum_{d_1=1}^{n-1} \cdots \sum_{d_r=1}^{n-1} \prod_{i=1}^r \left[ {\rm Pr}(d^{(i)}(u,v)=d_i)\right] \nonumber \\
  & \times & \mathbb{E}_{\Xi_{u,v}^{(1)},\ldots, \Xi_{u,v}^{(r)}}\left[g\bigl(\Xi_{u,v}^{(1)},\ldots,\Xi_{u,v}^{(r)}\bigr) \Bigm| d^{(i)}(u,v)=d_i, i \in [r] \right], \notag %
\end{IEEEeqnarray}
where
\begin{IEEEeqnarray*}{rCl}
\IEEEeqnarraymulticol{3}{l}{\mathbb{E}_{\Xi_{u,v}^{(1)},\ldots, \Xi_{u,v}^{(r)}}\left[g\bigl(\Xi_{u,v}^{(1)},\ldots,\Xi_{u,v}^{(r)}\bigr) \Bigm| d^{(i)}(u,v)=d_i, i \in [r] \right]}\\
&= &  \sum_{h_1=1}^{d_1} \cdots  \sum_{h_r=1}^{d_r} g(h_1,\ldots,h_r)  \\
&  & \times {d_1 \choose h_1} \cdots {d_r \choose h_r} p^{d_1+\cdots+d_r-(h_1+\cdots h_r)} (1-p)^{h_1+\cdots+h_r}.  \notag %
\end{IEEEeqnarray*}
Now, for a fixed pair of nodes $u,v$, $d^{(i)}(u,v) = 1$ with probability $\nicefrac{1}{(n-1)}$, $d^{(i)}(u,v) = 2$ with probability $(1-\nicefrac{1}{(n-1)}) \cdot \nicefrac{1}{(n-2)} = \nicefrac{1}{(n-1)}$, $d^{(i)}(u,v) = 3$ with probability $(1-\nicefrac{1}{(n-1)}) \cdot (1-\nicefrac{1}{(n-2)}) \cdot \nicefrac{1}{(n-3)} = \nicefrac{1}{(n-1)}$, etc. Hence, $d^{(i)}(u,v)$ follows a uniform distribution.
As a result,
\begin{IEEEeqnarray*}{rCl}
\IEEEeqnarraymulticol{3}{l}{\mathbb{E}_{\Xi_{u,v}^{(1)},\ldots, \Xi_{u,v}^{(r)}}\left[g\bigl(\Xi_{u,v}^{(1)},\ldots,\Xi_{u,v}^{(r)}\bigr) %
\right]}\\
  &=&\frac{1}{(n-1)^r}\sum_{d_1=1}^{n-1} \cdots \sum_{d_r=1}^{n-1} \sum_{h_1=1}^{d_1} \cdots \sum_{h_r=1}^{d_r} g(h_1,\ldots,h_r) \\
  &  &  \times {d_1 \choose h_1} \cdots {d_r \choose h_r} p^{d_1+\cdots+d_r-(h_1+\cdots h_r)} (1-p)^{h_1+\cdots+h_r}\\ \notag %
  &\overset{(a)}{\leq} & \frac{1}{(n-1)} \sum_{d=1}^{n-1} \sum_{h=1}^{d} g(h,\ldots,h)  {d \choose h}  p^{d-h} (1-p)^{h}
\end{IEEEeqnarray*}
which is independent of $u,v$, %
and where $(a)$ follows from the  fact that $g(\cdot,\ldots,\cdot)$ is a decreasing and convex function.   
Hence,
\begin{IEEEeqnarray}{rCl}
\IEEEeqnarraymulticol{3}{l}{\max_{\substack{u,v \in \set V,\\ \ u \neq v}} \RDiv[\alpha]{\set O^{(r)}_v(\set A(\set D))}{\set O^{(r)}_v(\set A(\set D'))}}\nonumber\\
&  &  \leq \frac{4 \alpha k^2}{(n-1)\sigma^2}\sum_{d=1}^{n-1} \sum_{h=1}^{d} g(h,\ldots,h) {d \choose h} p^{d-h} (1-p)^{h}. 
\label{eq:maxRenyiDivUpperBound}
\end{IEEEeqnarray}
As for the \ssring\ scheme, the number of visits of the token to a node $v$ during the execution of the algorithm, denoted by $\Xi_v$, follows a binomial distribution with parameters  $\nicefrac{h_{\max}}{n}$ and $1-p$. Let $\tilde{h}$ be defined as in the formulation of the theorem. Then, it follows from a standard  Chernoff bound that ${\mathrm{Pr}}(\Xi_v \geq \tilde{h}) \leq \delta'$, for some $\delta' \in (0,1)$. 
Applying the composition theorem for RDP (\cref{lem:summation-bounds_RDiv}), \cref{lem:RandomComp}, and swapping the order of maximization and summation as in the derivations in \eqref{eq:composition}, but using \eqref{eq:maxRenyiDivUpperBound} together with the definition of $g(\cdot)$ from \eqref{eq:maxRenyiDivRndRingDecLearnRate}, results in 
$\max_{u,v \in \set V,\ u \neq v} \RDiv[\alpha]{\set O_v(\set A(\set D))}{\set O_v(\set A(\set D'))} \leq  \frac{4 a \alpha k^2}{\sigma^2}$,
where $a$ is defined in the theorem formulation.

Then, converting from RDP to DP using \cref{lem:RDPtoDP}  gives that \cref{alg:ss} satisfies
\begin{IEEEeqnarray}{c} \label{eq:ring_all_noise_ndp}
\left( \frac{4  a \alpha k^2}{\sigma^2} + \frac{\log(\nicefrac{1}{\delta})}{\alpha-1}, \delta + \delta' \right)-\text{NDP},
\end{IEEEeqnarray}
where again the parameter $\alpha$ can be optimized in order to minimize the $\varepsilon$ (left) term  in \eqref{eq:ring_all_noise_ndp}. However, there is a subtlety as the condition in \eqref{eq:condition_alpha} must be satisfied. Taking the derivative of the $\varepsilon$ (left) term of \eqref{eq:ring_all_noise_ndp}  with respect to $\alpha$, equating it to zero, and setting $\sigma= \frac{k \sqrt{8 \log (\nicefrac{1.25}{\delta})}}{\varepsilon}$, where  $\varepsilon > 0$ 
and $0 < \delta < 1$, gives
 $\alpha =  1+ \frac{\sqrt{2\log (\nicefrac{1}{\delta}) \log (\nicefrac{1.25}{\delta})}}{\varepsilon \sqrt{ a}}
 > 1$  %
and the final result follows by substituting the minimum of the optimal value of $\alpha$ from above and the right-hand-side upper bound of \eqref{eq:condition_alpha} into the $\varepsilon$ (left) term of \eqref{eq:ring_all_noise_ndp} and simplifying.

\IEEEtriggeratref{46}
\bibliographystyle{IEEEtran}
\bibliography{\bibfiles/defshort1,\bibfiles/biblioHY,refs}
\ifCLASSOPTIONcaptionsoff
  \newpage
\fi

\ifthenelse{\boolean{MAIN_REF}}{}{\clearpage
\documentclass[privamp_jsait_final_arxiv.tex]{subfiles}

\ifSubfilesClassLoaded{%
  \setcounter{page}{1}
  \setcounter{theorem}{3}
  \setcounter{equation}{18}
  \setcounter{figure}{5}
}{}%

\newenvironment{Figure}
  {\par\medskip\noindent\minipage{\linewidth}}
  {\endminipage\par\medskip}

\title{Straggler-Resilient Differentially-Private\\[1mm] Decentralized  Learning: Supplementary Material}

\begin{document}

\twocolumn[%
\hrule height 4pt
\vskip 0.25in
\vskip -\parskip%
\centering
{\LARGE\bf Straggler-Resilient Differentially-Private\\[1mm] Decentralized  Learning: Supplementary Material\par}
\vskip 0.29in
\vskip -\parskip
\hrule height 1pt
\vskip 0.09in%
\vspace{2ex}
]

  \begin{bibunit}[IEEEtranSA]

\appendices

\setcounter{section}{2}

\section{Uniform Random Walk} \label{sec:uniform_random_walk}

For a uniform random walk on the set of nodes $\mathcal{V}$ the token uniformly at random picks an arbitrary node  in $\mathcal{V}$ with probability $\nicefrac{1}{n}$. If the node does not respond within the timeout period, the process is repeated until a node responds, i.e.,  the token uniformly at random picks an arbitrary node  in $\mathcal{V}$ with probability $\nicefrac{1}{n}$ until a response is received. Then, for \cref{alg:ss} and with a constant noise variance $\sigma_h=\sigma$, $\forall\, h$, the privacy leakage level $\varepsilon_{\mathrm{skip}}$  after $h_{\max}$ steps is given as follows.

\begin{theorem}
  \label{thm:ss_rand_walk_all_noise}
  Let $\varepsilon > 0$ 
  and $0 < \delta < 1$. Then, under Assumptions~\ref{ass:Lipschitz} and \ref{ass:smooth},
   a uniform random walk on a set of $n$ nodes %
  and with learning rate parameter  $0 < \zeta \leq \nicefrac{2}{\beta}$
   achieves $(\varepsilon_{\mathrm{skip}},\delta+\delta')$-NDP for all $\delta' \in (0,1]$ with
  \begin{IEEEeqnarray*}{c}
    \varepsilon_{\mathrm{skip}} =\frac{\varepsilon^2  a  \alpha}{2  \log (\nicefrac{1.25}{\delta})}   + \frac{\log (\nicefrac{1}{\delta})}{\alpha-1}, 
      \end{IEEEeqnarray*}
  where
  \begin{IEEEeqnarray*}{rCl}
  a& \triangleq & \frac{2}{n} \sum_{r=0}^{\tilde{h}-1} \sum_{d=1}^{\infty} \left(1-\frac{2}{d}\right)^{d-1} \sum_{h=1}^{d} \frac{h{d \choose h} p^{d-h} (1-p)^{h}}{ \gamma_{r,h} }, \\
            \gamma_{r,h}  & \triangleq  & 4 (1 +r \cdot h) \cdot \left(\sqrt{1 + r \cdot h + h} 
 - \sqrt{1 + r \cdot h}\right)^2,\\
 \tilde{h} &\triangleq &\left\lceil \nicefrac{h_{\max}(1-p)}{n}+ \sqrt{\nicefrac{3h_{\max}(1-p)}{n}   \log \left( {\nicefrac{1}{\delta'}} \right)} \right\rceil,
    \\
    \alpha& \triangleq &\Scale[1.0]{\min\left( \frac{2\sqrt{   \log (\nicefrac{1}{\delta}) \log (\nicefrac{1.25}{\delta})}}{\varepsilon \sqrt{a}} + 1,\frac{1+\sqrt{\frac{16  \log \left(\nicefrac{1.25}{\delta}\right)}{\varepsilon^2} +1 }}{2} \right)},\IEEEeqnarraynumspace
  \end{IEEEeqnarray*}
  and $0 \leq p < 1$ is the probability of skipping a node.
\end{theorem}

Note that the formulations in \cref{thm:ss_rand_ring_all_noise,thm:ss_rand_walk_all_noise} are similar, except for the definition of the parameter $a$.  %

\begin{figure}[t]
\begin{center}
\includegraphics[width=0.75\columnwidth]{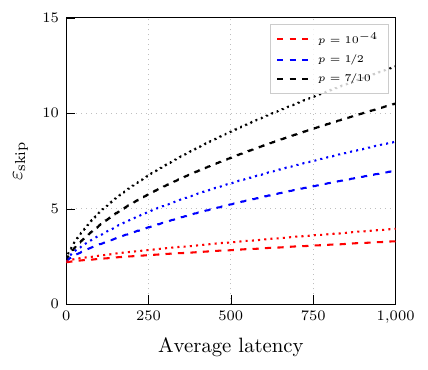}
\end{center}
\vspace{-3ex}
\caption{Privacy leakage level $\varepsilon_{\mathrm{skip}}$ vs average latency for the gamma delay model with shape $1/4$ and scale $1$,  and with the same parameters as in \cref{sec:numerical_results_theory} ($\delta'=10^{-6}$, $\epsilon=1$, $\delta=10^{-6}$, and $\chi=\nicefrac{1}{100}$), but with $n=100$ nodes. Dashed lines are for the \ssrandring\ scheme (\cref{thm:ss_rand_ring_all_noise}), while the dotted lines are for a uniform random walk on the set of nodes $\mathcal{V}$ (\cref{thm:ss_rand_walk_all_noise}).} \label{fig:RandomRingVersusRandomWalk}
\end{figure}

\begin{IEEEproof}
The proof follows analogous to the proof of \cref{thm:ss_rand_ring_all_noise}. However, due to the uniform random walk on the set of nodes $\set V$, $d^{(i)}(u,v)$ follows a geometric distribution with success probability $\nicefrac{2}{n}$ for any pair $(u,v)$, while for the \ssrandring\ scheme $d^{(i)}(u,v)$ follows a uniform distribution.  %
Following analogous derivations as in the proof of \cref{thm:ss_rand_ring_all_noise}, gives
\begin{IEEEeqnarray*}{rCl}
\IEEEeqnarraymulticol{3}{l}{\mathbb{E}_{\Xi_{u,v}^{(1)},\ldots, \Xi_{u,v}^{(r)}}\left[g\bigl(\Xi_{u,v}^{(1)},\ldots,\Xi_{u,v}^{(r)}\bigr) %
\right]}\\
  &=&\frac{2^r}{n^r}\sum_{d_1=1}^{\infty} \!\cdots \!\!\sum_{d_r=1}^{\infty} \left(1-\frac{2}{n} \right)^{d_1+\cdots+d_r-r} \\
  && \quad \times \sum_{h_1=1}^{d_1}\!\cdots \!\!\sum_{h_r=1}^{d_r} g(h_1,\ldots,h_r) %
   {d_1 \choose h_1} \!\cdots \! {d_r \choose h_r} \\
   && \quad \times p^{d_1+\cdots+d_r-(h_1+\cdots h_r)} (1-p)^{h_1+\cdots+h_r}\\ \notag %
  &\overset{(a)}{\leq} & \frac{2}{n} \sum_{d=1}^{\infty}  \left(1-\frac{2}{n} \right)^{d-1} \sum_{h=1}^{d} g(h,\ldots,h)  {d \choose h}  p^{d-h} (1-p)^{h},
\end{IEEEeqnarray*}
where $(a)$ follows from the  fact that $g(\cdot,\ldots,\cdot)$ is a decreasing and convex function.   
The analogue of \eqref{eq:maxRenyiDivUpperBound} becomes
\begin{IEEEeqnarray*}{rCl}
\IEEEeqnarraymulticol{3}{l}{\max_{\substack{u,v \in \set V,\\ \ u \neq v}} \RDiv[\alpha]{\set O^{(r)}_v(\set A(\set D))}{\set O^{(r)}_v(\set A(\set D'))}}\nonumber\\
&  &  \leq \frac{8 \alpha k^2}{n\sigma^2}\sum_{d=1}^{\infty} \left(1-\frac{2}{n}\right)^{d-1} \sum_{h=1}^{d} g(h,\ldots,h) {d \choose h} p^{d-h} (1-p)^{h}, \nonumber \\*
\label{eq:maxRenyiDivUpperBound_randomwalk}
\end{IEEEeqnarray*}
from which the result follows in a similar manner as in the rest of the proof of \cref{thm:ss_rand_ring_all_noise} from below \eqref{eq:maxRenyiDivUpperBound}.
\end{IEEEproof}

Due to the geometric distribution for $d^{(i)}(u,v)$ (see proof above), the privacy leakage level $\varepsilon_{\mathrm{skip}}$ is higher than for the \ssrandring\ scheme as the probability of a small value of $d^{(i)}(u,v)$ is higher. For instance, $d^{(i)}(u,v)=1$ with probability $2/n$ for a uniform random walk on   a set of $n$ nodes, while for the \ssrandring\ scheme this probability is $1/(n-1)$, which is about half smaller.

In \cref{fig:RandomRingVersusRandomWalk}, we plot  privacy leakage level $\varepsilon_{\mathrm{skip}}$ versus  average latency for the gamma delay model with shape $1/4$ and scale $1$, and  with the same parameters as in \cref{sec:numerical_results_theory} ($\delta'=10^{-6}$, $\epsilon=1$, $\delta=10^{-6}$, and $\chi=\nicefrac{1}{100}$), but with $n=100$ nodes. Dashed lines are for the \ssrandring\ scheme (\cref{thm:ss_rand_ring_all_noise}), while the dotted lines are for a uniform random walk on the set of nodes $\mathcal{V}$ (\cref{thm:ss_rand_walk_all_noise}). As can be seen from the figure, the privacy leakage level $\varepsilon_{\mathrm{skip}}$ with the \ssrandring\ scheme is lower than with a uniform random walk.
\section{Proofs} \label{app:proofs}

\subsection{Proof of \cref{rem:1}}

The result for the \ssring\   scheme follows directly from the expression for $\varepsilon_{\mathrm{skip}}$ given in  \cref{thm:ss_ring_all_noise} as $\tilde{h}$ is linear in $h_{\max}$, while  for \cref{thm:ss_rand_ring_all_noise} it follows by lower-bounding $\gamma_{r,h}$ by $\gamma_{0,h}$ for all $0 \leq r \leq \tilde{h}-1$ as $\gamma_{r,h}$ is a strictly increasing function in $r$ for a fixed $h > 0$ (see the proof in the next paragraph below). Then, $a$ is of order $O(h_{\max})$ as 1) the double inner summation in the expression for $a$ can be upper-bounded by a constant, and  2) $\tilde{h}$ is linear in $h_{\max}$. Moreover, the value of $\alpha$ will approach $1$ as $h_{\max}$ increases as $a$ is of order $O(h_{\max})$ (the first term in the minimization for $\alpha$ gives the minimum).  The second term in the expression for $\varepsilon_{\mathrm{skip}}$ is of order $O(\sqrt{h_{\max}})$ as $\alpha-1$ is of order $O(1/\sqrt{h_{\max}})$, while the first term is of order $O(h_{\max})$ ($\alpha$ approaches $1$),  from which the result follows.

Next, we proceed to the second part of the proof, showing that the function
\begin{align*}
f(r;h) = (1 +r \cdot h) \cdot \left(\sqrt{1 + r \cdot h + h}  - \sqrt{1 + r \cdot h}\right)^2
\end{align*}
is strictly increasing in $r$ for any fixed $h>0$. This follows by showing that the derivative with respect to $r$ is strictly positive. By the chain rule, we get
\begin{IEEEeqnarray*}{rCl}
	\frac{\partial f}{\partial r} 
	 & = & h \left( \sqrt{1 + r \cdot h + h}  - \sqrt{1 + r \cdot h} \right)^2\\
	 &&+\;(1 +r \cdot h) \left( \sqrt{1 + r \cdot h + h}  - \sqrt{1 + r \cdot h} \right) \\
	 && \times \left( \frac{h}{\sqrt{1 + r \cdot h + h}} - \frac{h}{\sqrt{1 + r \cdot h}}\right)\\
	 &=& h \left( \sqrt{1 + r \cdot h + h}  - \sqrt{1 + r \cdot h} \right) \\
	 && \times \left( \sqrt{1 + r \cdot h + h} - 2 \sqrt{1 + r \cdot h} +   \frac{1 +r \cdot h}{\sqrt{1 + r \cdot h + h}} \right) \\
	 &=& \frac{h \left( \sqrt{1 + r \cdot h + h}  - \sqrt{1 + r \cdot h} \right)^3}{ \sqrt{1 + r \cdot h + h}} > 0.
\end{IEEEeqnarray*}
Hence, $\gamma_{r,h} \geq \gamma_{0,h}$.

\subsection{Proof of \cref{prop:latency}}
The time between two consecutive nodes (either straggling or not) consists of the constant communication latency, $\chi$, and the random waiting time in a node, which has the expected value
	\begin{align*}
		\EE{\min(T, \tss)} & = \int_{0}^{\infty} \min(t, \tss) \mathrm d \Phi_T(t) \\
		&= \int_{0}^{\tss} t \mathrm d \Phi_T(t) + \tss \int_{\tss}^{\infty} \mathrm d \Phi_T(t) \\
		&= \int_{0}^{\tss} t \mathrm d \Phi_T(t) + \tss (1 - \Phi_T(\tss))
	\end{align*}
	and the result follows.

\subsection{Proof of \cref{prop:optimal_tss}}
	Denote by $Z \geq 0$ the number of stragglers between two consecutive nonstraggling nodes. It follows the geometric distribution with success probability $1-p = \Phi_T(\tss)$. Then, the average latency between two token updates is
		\begin{multline*}
		\EE{Z} (\chi + \tss) + \chi + \EE{T \mid T \le \tss} \\
		= \frac{1-\Phi_T(\tss)}{\Phi_T(\tss)} (\chi + \tss) + \chi + \int_{0}^{\tss} t \mathrm d \frac{\Phi_T(t)}{\Phi_T(\tss)} \\
		= \frac{\chi + \int_{0}^{\tss} t \mathrm d \Phi_T(t) + \tss (1-\Phi_T(\tss))}{\Phi_T(\tss)}.
	\end{multline*}
	To get the minimal average latency between token updates, one only needs to minimize this expression in $\tss$.

\subsection{Proof of \cref{lem:markov-mixing}}

	Since the state transition probability matrix $Q$ is circulant (cf. \cite{tee2007eigenvectors}), its eigenvalues $\lambda_j$ and corresponding eigenvectors $\varphi_j$ are
	\begin{IEEEeqnarray*}{rCl}
		\lambda_j &=& \frac{(1-p)\omega^j}{1-p \omega^j}, \quad j \in [0:n-1], \\
		\varphi_j &=& \frac{1}{\sqrt n} (1, \omega^j, \omega^{2j}, \dotsc, \omega^{(n-1)j})^\top, \quad j \in [0:n-1],
	\end{IEEEeqnarray*}
	where $\omega = \mathrm e^{\nicefrac{2 \pi \mathfrak i}{n}}$ is a primitive $n$-th root of unity and $\mathfrak i = \sqrt{-1}$ is the imaginary unit. The absolute values of the eigenvalues are
	\[
	|\lambda_j| = \frac{1-p}{\sqrt{1+p^2 - 2 p \cos \frac{2 \pi j}n}},  \quad j \in [0:n-1],
	\]
	and $|\lambda_j| \le |\lambda_1| < \lambda_0=1$ for $j  \in [n-1]$. Let $U$ be a matrix whose columns are $\varphi_0,  \varphi_1, \dotsc,  \varphi_{n-1}$. The matrix $U$ is unitary (i.e., $U^* = U^{-1}$) and we can diagonalize $Q$ as 
		\[
	Q = U \diag(1, \lambda_1, \dotsc, \lambda_{n-1}) U^*.
	\]
	Next,
	\begin{IEEEeqnarray*}{rCl}
		\lim_{h \to \infty}	Q^h &=& \lim_{h \to \infty} U \diag(1, \lambda_1^h, \dotsc, \lambda_{n-1}^h) U^* \\
		&=& U \diag(1, 0, \dotsc, 0) U^*
                \\[1mm]
		&=& \begin{pmatrix}
                  \frac 1n & \frac 1n & \cdots & \frac 1n
                  \\[1mm]
                  \frac 1n & \frac 1n & \cdots & \frac 1n
                  \\[1mm]
                  \vdots & \vdots & \ddots & \vdots
                  \\
                  \frac 1n & \frac 1n & \cdots & \frac 1n 
		\end{pmatrix} \eqdef Q^\infty.
	\end{IEEEeqnarray*}
	Finally,
        \begin{IEEEeqnarray*}{rCl}
          \norm[1]{ \pi^{(h)} -  \pi^{(\infty)}} & \stackrel{(a)}{\le} &\sqrt{n} \norm[2]{ \pi^{(h)} -  \pi^{(\infty)}}
          \\
          & \stackrel{(b)}{=} &\sqrt n \norm[2]{(Q^{h})^\top  \pi^{(0)} - (Q^\infty)^\top  \pi^{(0)}}
          \\
          & \stackrel{(c)}{\le} &\sqrt n \norm[2]{(Q^h)^\top - (Q^\infty)^\top} \norm[2]{ \pi^{(0)}}
          \\
          & \stackrel{(d)}{\le} &\sqrt n \norm[2]{(Q^h)^\top - (Q^\infty)^\top}
          \\
          & \stackrel{(e)}{=} &\sqrt n \norm[2]{Q^h - Q^\infty}
          \\		
          & \stackrel{(f)}{=} &\sqrt n \norm[2]{U \diag(0, \lambda_1^h, \lambda_2^h, \dotsc, \lambda_{n-1}^h) U^*}
          \\
          & \stackrel{(g)}{=} &\sqrt n \norm[2]{\diag(0, \lambda_1^h, \lambda_2^h, \dotsc, \lambda_{n-1}^h)}
          \\
          & \stackrel{(h)}{=} &\sqrt{n} |\lambda_1|^h.
        \end{IEEEeqnarray*}
	Here, 
	$(a)$ is from a general relation between the $\ell_1$ and $\ell_2$ norms in $\Reals^n$; 
	$(b)$ is a standard expression for Markov chains; 
	$(c)$ is by the definition of induced norm;\footnote{Note that the matrix norm $\norm[2]{\cdot}$ induced by the vector norm $\norm[2]{\cdot}$ is the \emph{spectral norm}, which is different from the entry-wise norm.} 
	$(d)$ is because $\norm[2]{ \pi^{(0)}} \le \norm[1]{ \pi^{(0)}} = 1$ as $\pi^{(0)}$ is a probability vector; 
	$(e)$ is because the spectral norm is invariant under transposition; 
	$(f)$ is a simple matrix manipulation; 
	$(g)$ is because multiplying by a unitary matrix does not change the spectral norm; 
	and $(h)$ is direct calculation of the spectral norm.

\subsection{Proof of  \cref{lem:moments-of-gaussian-norm}}

  The random variable $X = \frac 1{\sigma} \norm[2]{N}$ is distributed according to a $\chi$ distribution with $d$ degrees of freedom. Then
  \[
  	\EE{\norm[2]{N}} = \sigma \EE{X} = \sigma \sqrt 2 \frac{ \Gamma\left( \frac{d+1}2\right)}{\Gamma\left(\frac{d}2 \right)} 
  	\overset{(a)}{<} \sigma \sqrt{d},
  \]
  where $\Gamma(\cdot)$ denotes the gamma function and $(a)$  follows from the logarithmic convexity of $\Gamma$. Also,
  \[
    \EE{\norm[2]{N}^2} = \sigma^2 \EE{X^2} = 2 \sigma^2 \frac{ \Gamma\left( \frac{d}2 +1\right)}{\Gamma\left(\frac{d}2 \right)} = d \sigma^2.
  \]
  
  \subsection{Proof of \cref{lem:RandomComp}}

  Denote the $i$-fold composition $(\set{A}_1,\ldots,\set{A}_i)$ by $\set{A}^{(i)}$, and its privacy guarantee by $(\veps^{(i)}_{\mathrm{c}},\delta^{(i)}_\mathrm{c})$, $i\in [r]$. %
 For any set $\set{S}$, we have
  \begin{IEEEeqnarray*}{rCl}
    \IEEEeqnarraymulticol{3}{l}{%
      \Prv{\set{A}^{(R)}(\set{D})\in\set{S}}}\nonumber\\*\,%
    & = &\Prv{\{\set{A}^{(R)}(\set{D})\in\set{S}\}\cap\{R\leq r\}}\nonumber\\
    && +\>\underbrace{\Prv{\{\set{A}^{(R)}(\set{D})\in\set{S}\}\cap\{R> r\}}}_{\leq\Prs{R> r}}
    \\
    & \leq &\sum_{i=1}^r\bigPrv{R=i}\Prvcond{\set{A}^{(R)}(\set{D})\in\set{S}}{R=i}+\Prs{R> r}
    \\
    & \stackrel{(a)}{\leq} &\sum_{i=1}^r\Prv{R=i}\Bigl[\ee^{\veps_\mathrm{c}^{(i)}}\bigPrvcond{\set{A}^{(R)}(\set{D}')\in\set{S}}{R=i}+\delta^{(i)}_\mathrm{c}\Bigr]\nonumber\\
    && +\>\Prs{R> r}
    \\
    & \stackrel{(b)}{\leq} &\sum_{i=1}^r\Prv{R=i}\Bigl[\ee^{\veps_\mathrm{c}^{(r)}}\bigPrvcond{\set{A}^{(R)}(\set{D}')\in\set{S}}{R=i}+\delta^{(r)}_\mathrm{c}\Bigr]\nonumber\\
    && +\>\delta'
    \\
    & \leq &\ee^{\veps_\mathrm{c}^{(r)}}\bigPrv{\{\set{A}^{(R)}(\set{D}')\in\set{S}\}\cap\{R\leq r\}}\nonumber\\
    && +\>\delta^{(r)}_\mathrm{c}\underbrace{\Prs{R\leq r}}_{\leq 1} + \delta'
    \\
    & \leq &\ee^{\veps_\mathrm{c}^{(r)}}\bigPrv{\set{A}^{(R)}(\set{D}')\in\set{S}}+\delta^{(r)}_\mathrm{c}+\delta',\IEEEeqnarraynumspace
  \end{IEEEeqnarray*}
  where $(a)$ holds by applying the $i$-fold composition theorem for DP, and $(b)$ follows because $(\veps^{(i)}_{\mathrm{c}},\delta^{(i)}_{\mathrm{c}})$ is monotonically increasing in $i\in[r]$. This completes the proof.

\subsection{Proof of \cref{lem:technical_xi}}

From \eqref{eq:varrho_uv},
  \begin{IEEEeqnarray*}{rCl} 
\left(\varrho_{u,v}^{(r+1)}\right)^2 & = &
\left(\sum_{i \in [0:\ell-c]} \sqrt{\frac{h_c}{h_c+i}} \right)^2 \\
 & = &\sum_{i \in [0:\ell-c]} \frac{h_c}{h_c+i} + 2 \sum_{i=0}^{\ell -c} \sum_{j=0}^{i-1}   \sqrt{\frac{h_c}{h_c+i}} \sqrt{\frac{h_c}{h_c+j}}\\
  & \geq &\sum_{i \in [0:\ell-c]} \frac{h_c}{h_c+i} + 2 \sum_{i=0}^{\ell -c}    i \frac{h_c}{h_c+i} \\
  & = & \sum_{i \in [0:\ell-c]}  \frac{h_c+ 2i h_c}{h_c+i} \geq \ell-c+1
  \end{IEEEeqnarray*}
  which completes the proof.

\begin{table*}[t]
	\centering
	\caption{Neural network architectures used for training for the MNIST and CIFAR-$10$ image datasets. The input pixel value is rescaled between $-1.0$ and $1.0$. FC stands for a fully connected layer. Conv  stands for a   convolutional layer and ``st'' is shorthand for stride. Size is the number of neurons or input size. As activation function we used SeLU and softmax.} \label{tab:table_NNarch} %
\vspace{-2ex}
	\begin{tabular}{@{}lp{0.8\textwidth}@{}}
		\toprule
		\multicolumn{2}{c}{Detailed architecture (consecutive layers' sizes and types)} \\
		\midrule
		MNIST
		& $1 \times 28 \times 28 \times 1$ (Input), $1 \times 26 \times 26 \times 64$ (Conv, SeLU), 
		                                                   $1 \times 12 \times 12 \times 64$ (Conv ($\textnormal{st}=2$), SeLU), $1 \times 10 \times 10 \times 128$ (Conv, SeLU),
		                                                   $1 \times 4 \times 4 \times 128$ (Conv ($\textnormal{st}=2$), SeLU), $10$ (FC, softmax)
		\\                                                   %
		\midrule
		CIFAR-$10$ &
		$1 \times 32 \times 32 \times 3$ (Input), $1 \times 32 \times 32 \times 64$ (Conv, SeLU), 
		         $1 \times 16 \times 16 \times 64$ (Conv ($\textnormal{st}=2$), SeLU), $1 \times 16 \times 16 \times 128$ (Conv, SeLU), $1 \times 8 \times 8 \times 128$ (Conv ($\textnormal{st}=2$), SeLU), $1 \times 8 \times 8 \times 256$ (Conv, SeLU), $1 \times 8 \times 8 \times 256$ (Conv, SeLU), $1 \times 4 \times 4 \times 256$ (Conv ($\textnormal{st}=2$), SeLU), $1 \times 4 \times 4 \times 512$ (Conv, SeLU), $1 \times 4 \times 4 \times 512$ (Conv, SeLU), $1 \times 2 \times 2 \times 512$ (Conv ($\textnormal{st}=2$), SeLU), $10$ (FC, softmax)
		\\
		\bottomrule
	\end{tabular}
\end{table*}
\section{Asymptotic Convergence Rate} 
\label{sec:asympt-conv-rate}

Define a \emph{positive binomial RV}
$B$ with parameters $n = h_{\max}$ and success probability $q=1-p$ by
\begin{IEEEeqnarray*}{c}
  \Prv{B=h}=\frac{1}{1-p^n}\binom{n}{h}q^h(1-q)^{n-h},\quad h\in[1:n].
\end{IEEEeqnarray*}
We are interested in the asymptotic behavior of
\begin{IEEEeqnarray*}{c}
  (1-p^n)\E{\frac{\log{B}}{\sqrt{B}}}=\sum_{h=1}^n\frac{\log{h}}{\sqrt{h}}\binom{n}{h}q^h(1-q)^{n-h}.
\end{IEEEeqnarray*}
Since $p$ is fixed, we can focus on the quantity $\E{\frac{\log{B}}{\sqrt{B}}}$. 
We first use the following inequality for the logarithm function, which simply states that the concave function $\log(x)$ lies below its tangent line at a point $a > 0$.\footnote{For a formal proof, see, e.g., \cite[Lem.~2.29]{Yeung08_1} for the case of $a=1$. The general case for an arbitrary $a>0$ can be proved in the same way.}
\begin{proposition}
  \label{prop:upper-bound_log-ft}
  For any $a>0$ and $x>0$, we have
  \begin{IEEEeqnarray*}{c}
    \log{x}\leq\frac{x}{a}+\log{\frac{a}{\mathrm{e}}}.
  \end{IEEEeqnarray*}
\end{proposition}%
Hence, we have %
\[
	\E{\frac{\log{B}}{\sqrt{B}}} \leq \E{\frac{\frac{B}{a}+\log\frac{a}{\mathrm{e}}}{\sqrt{B}}}
	 = \frac{\E{\sqrt{B}}}{a}+\E{\frac{1}{\sqrt{B}}}\log\frac{a}{\mathrm{e}},
\]
and, with $a \triangleq \frac{\E{\sqrt{B}}}{\E{\frac{1}{\sqrt{B}}}}>0$, it becomes
\begin{IEEEeqnarray*}{rCl}
	\E{\frac{\log{B}}{\sqrt{B}}}& \leq &\frac{\BigE{\sqrt{B}}}{\frac{\bigE{\sqrt{B}}}{\E{\frac{1}{\sqrt{B}}}}}+\E{\frac{1}{\sqrt{B}}}\log\frac{\frac{\bigE{\sqrt{B}}}{\E{\frac{1}{\sqrt{B}}}}}{\mathrm{e}}
	\\
	& = &\E{\frac{1}{\sqrt{B}}}+\E{\frac{1}{\sqrt{B}}}\log{\frac{\BigE{\sqrt{B}}}{\BigE{\frac{1}{\sqrt{B}}}}}\nonumber\\
	&& -\>\E{\frac{1}{\sqrt{B}}}\log{\mathrm{e}}
	\\
	& = &\E{\frac{1}{\sqrt{B}}}\log{\frac{\BigE{\sqrt{B}}}{\BigE{\frac{1}{\sqrt{B}}}}}.\IEEEyesnumber\label{eq:2nd-upper-bound_expected-objective-ft}
\end{IEEEeqnarray*}%
Since the function $f=\sqrt{x}$ is concave and $g=\frac{1}{\sqrt{x}}$ is convex for $x>0$, we can use  Jensen's inequality~\cite[Ch.~2.6]{CoverThomas06_1} and get
\begin{IEEEeqnarray*}{rCl}
	\BigE{\sqrt{B}}& \leq &\sqrt{\E{B}}=\sqrt{\textstyle\frac{1}{1-p^n}\sum_{h=1}^n h\binom{n}{h}q^h(1-q)^{n-h}}
	\\
	& = &\sqrt{\frac{nq}{1-p^n}} \quad \text{and}
	\\
	\biggE{\frac{1}{\sqrt{B}}}& \geq &\frac{1}{\sqrt{\E{B}}}=\frac{1}{\sqrt{\frac{nq}{1-p^n}}}.
\end{IEEEeqnarray*}
Hence, \eqref{eq:2nd-upper-bound_expected-objective-ft} can be further bounded from above by
\begin{IEEEeqnarray*}{rCl}
  \E{\frac{\log{B}}{\sqrt{B}}}& \leq &\E{\frac{1}{\sqrt{B}}}\log{O(n)}.
\end{IEEEeqnarray*}
Finally, we use the asymptotic result from \cite[Thm.~1]{Znidaric09_1}, which indicates that
\[
	\E{\frac{1}{\sqrt{B}}} = O\left(\frac{nq}{(nq+p)^{1+\frac{1}{2}}}\right) = O\left(\frac{1}{\sqrt{n}}\right).
\]
\section{Neural Network Architectures}

In this appendix, the architectures of the deep neural networks used for training for the MNIST and CIFAR-$10$ datasets are detailed (see \cref{tab:table_NNarch}). For all three considered computation latency models, the same neural network architecture was used with the same initial learning rate.

\section{Latency Calculation} \label{app:additional_num_results}

In this appendix, we present %
a more detailed description of the latency calculations for \texttt{Muffliato-SGD} and \texttt{FedL-SGD}. We note, however, that the setups of \ssrandring, \texttt{Muffliato-SGD}, and \texttt{FedL-SGD} are very different and that it is challenging to make the latency comparison completely fair.

\subsection{\texttt{Muffliato-SGD}}

For \texttt{Muffliato-SGD}, we simulate skipping a node with probability $p$ by reducing the connectivity of the random  Erd\H{o}s-R{\'e}nyi communication graph generated in each round of gossiping. In particular,  as each node has degree $\log n$ (as  in the setup of \cite[Fig.1(c)]{CyffersEvenBelletMassoulie22_1}), we consider random Erd\H{o}s-R{\'e}nyi graphs with nodes of degree $(1-p) \log n$.
Next, the overall latency is simulated by first generating $\log n$ independent realizations of a random variable distributed according to the considered delay model (exponential, gamma, or Pareto type II) and then taking the minimum of $t_{\textnormal{skip}}$ (for the considered value of $p$; see Appendix~\ref{app:tables}) and the maximum of these  $\log n$ realizations, denoted by $t_{\textnormal{comp}}$ in the following. Second, we add the average communication latency of each round of gossiping according to the random access scheme in \cite{ChenDahlLarsson23_1}  with an optimized value for the probabilistic random access policy, which is equivalent to adding $\chi$ (the latency of a single broadcast transmission from each node) times the average number of time slots needed  for each node to successfully send a message to all of its neighbors using the random access procedure. The probabilistic random access scheme with broadcast transmission works in the following way. In each time slot, each node  independently broadcasts its gradient with probability $q$ (to be optimized and referred to as the probabilistic random access policy). In each time slot, a node can receive at most one gradient successfully from its neighbors. When multiple nodes broadcast to a common neighbor, it will result in a collision and no information will be successfully received. Hence,  a node $u$ can successfully receive information from its neighbor node $v$ if and only if: 1) node $v$ decides to broadcast; 2) node $u$ decides not to broadcast; and 3) all neighbors of node $u$ except node $v$ decide not to broadcast. The procedure  is repeated until all nodes have successfully transmitted their gradients to all neighbors. For random Erd\H{o}s-R{\'e}nyi graphs with $n=1000$ nodes and a node degree of $\log n$ the number of random access repetitions is approximately $175$ on average with an optimized random access policy of $q=\nicefrac{15}{100}$. Hence, the average overall  latency due to gossiping becomes $t_{\textnormal{comp}} + T_\textnormal{gossip} \cdot 175 \cdot \chi$, where $T_\textnormal{gossip}$ is the number of gossip iterations.

\subsection{\texttt{FedL-SGD}}
For \texttt{FedL-SGD}, we simulate skipping a node with probability $p$ by first generating $n$ independent realizations of a random variable distributed according to  the considered delay model (exponential, gamma, or Pareto type II). Next, the  latency for each iteration of upload from the $n$ nodes to the central server and download from the central server to the $n$ nodes is calculated 
by first taking the minimum of $t_{\textnormal{skip}}$ (for the considered value of $p$; see Appendix~\ref{app:tables}) and the maximum of these  $n$ realizations, denoted by $t_{\textnormal{comp}}$ in the following. Second, we add the  communication latency of each iteration round. For the upload this is $\chi$ times the number of nodes $n$  divided by the number of independent subchannels $N_\textnormal{c}$. For the download, we assume a single broadcast, giving a latency contribution of $\chi$. Hence, the  overall  latency for each iteration becomes $t_{\textnormal{comp}} +  (1+ \nicefrac{n}{N_\textnormal{c}}) \chi$.

\section{Computation Latency  Models} \label{app:tables}

The following computation latency models  were used throughout this work ($\Phi_T'(t)$ denotes the derivative of $\Phi_T(t)$).

\begin{enumerate}
\item Exponential with mean $1$:
\begin{align*}
	\Phi_T(t)  &= 1-{\mathrm e}^{-t}, \\
	\Phi_T'(t) &= {\mathrm e}^{-t}.
\end{align*}

\item Gamma with shape $\nicefrac{1}{4}$ and scale $1$:
\begin{align*}
	\Phi_T(t)  &= \frac{1}{\Gamma(\nicefrac 14)} \int_{0}^t x^{-\nicefrac 34} {\mathrm e}^{-x} \mathrm{d} x, \\
	\Phi_T'(t) &= \frac{t^{-\nicefrac 34} {\mathrm e}^{-t}}{\Gamma(\nicefrac 14)}.
\end{align*}

\item Pareto type II (also known as Lomax) with shape $3$ and scale $2$:
\begin{align*}
	\Phi_T(t)  &= 1-\frac{8}{(t+2)^3}, \quad t \ge 0,\\
	\Phi_T'(t) &= \frac{24}{(t+2)^4}, \quad t \ge 0.
\end{align*}
\end{enumerate}

Using these and \cref{prop:optimal_tss}, we  get the \emph{best} values of $\tss$ and $p$ as follows.

\begin{center}
\begin{tabular}{@{}lll@{}}
\toprule
Distribution   &   Best $\tss$  &     Best $p$ \\
\midrule
Exponential    &    $+\infty$   &         0
\\
Gamma          &    $0.00480$    &  $0.70986$ \\
Pareto  type II       &    $0.21390$   &      $0.73726$ \\
\bottomrule
\end{tabular}
\end{center}

We use the following values for $p$ and $\tss$ in \cref{fig:EPSandERRvsLAT,fig:EPSvsERR,fig:Accuracy_empirical,fig:Muffliato_comparisons,fig:RandomRingVersusRandomWalk}.
\begin{center}
  \begin{tabular}{@{}llll@{}}
    \toprule
       && $\tss$ \\
    \cmidrule{2-4}
     & Exponential & Gamma & Pareto type II\\
    \midrule
    $p=10^{-4}$ & 9.21034 & 6.42831   & 41.0887\\
    $p=\nicefrac 12$     & 0.69315 & 0.04367 & 0.51984\\
    $p=\nicefrac{7}{10}$     & 0.35667 & 0.00549 & 0.25250\\
    \bottomrule
  \end{tabular}
\end{center}%

\ifCLASSOPTIONcaptionsoff
  \newpage
\fi

\balance
\putbib[\bibfiles/defshort1,\bibfiles/biblioHY,refs]

\end{bibunit}

\end{document}
}

\end{document}